\documentclass[preprint,12pt]{elsarticle}

\usepackage{graphicx}
\usepackage{epsfig}
\usepackage{multirow}
\usepackage[ruled,vlined]{algorithm2e}
\usepackage{color}
\usepackage{xparse}
\usepackage{xspace}
\usepackage{url}
\usepackage{booktabs}
\usepackage{amssymb}
\def\cl@chapter{\@elt {theorem}}
\makeatother
\usepackage{amsmath,amsfonts}
\usepackage[title]{appendix}%
\usepackage{xcolor}%
\usepackage{textcomp}%
\usepackage{manyfoot}%
\usepackage{booktabs}%
\usepackage{algpseudocode}%
\usepackage{listings}%
\usepackage{array}
\usepackage[caption=false,font=normalsize,labelfont=sf,textfont=sf]{subfig}
\usepackage{stfloats}
\usepackage{verbatim}
\usepackage{makecell}
\usepackage{bm}
\usepackage{microtype}
\usepackage{hhline}
\usepackage{tabularx}
\usepackage{hyperref}
\usepackage[capitalize]{cleveref}
\crefname{figure}{figure}{figures}
\Crefname{figure}{Figure}{Figures}
\crefname{section}{section}{sections}
\Crefname{section}{Section}{Sections}
\crefname{table}{table}{tables}
\Crefname{table}{Table}{Tables}
\crefname{equation}{}{}
\Crefname{equation}{Equation}{Equations}
\definecolor{orange}{rgb}{1.0, 0.5, 0.0}

\RequirePackage{xspace}
\makeatletter
\DeclareRobustCommand\onedot{\futurelet\@let@token\@onedot}
\def\@onedot{\ifx\@let@token.\else.\null\fi\xspace}
 
\def\ie{\emph{i.e}\onedot}

\makeatother
\newlength{\Oldarrayrulewidth}

\journal{Pattern Recognition}

\begin{document}

\begin{frontmatter}



\title{Scene Structure Guidance Network: Unfolding Graph Partitioning into Pixel-Wise Feature Learning}


\author[a]{Jisu Shin}
\author[a]{Seunghyun Shin}
\author[a]{Hae-Gon Jeon\corref{cor1}}
\ead{haegonj@gist.ac.kr}
\cortext[cor1]{Corresponding Author}

\affiliation[a]{organization={AI Graduate School, GIST},
            addressline={123, Cheomdangwagi-ro}, 
            city={Buk-Gu},
            postcode={61005}, 
            state={Gwangju},
            country={Korea}}

\begin{abstract}
Understanding the informative structures of scenes is essential for low-level vision tasks. Unfortunately, it is difficult to obtain a concrete visual definition of the informative structures because influences of visual features are task-specific.
In this paper, we propose a single general neural network architecture for extracting task-specific structure guidance for scenes.
To do this, we first analyze traditional spectral clustering methods, which computes a set of eigenvectors to model a segmented graph forming small compact structures on image domains. We then unfold the traditional graph-partitioning problem into a learnable network, named \textit{Scene Structure Guidance Network (SSGNet)}, to represent the task-specific informative structures.   
The SSGNet yields a set of coefficients of eigenvectors that produces explicit feature representations of image structures. In addition, our SSGNet is light-weight ($\sim$ 56K parameters), and can be used as a plug-and-play module for off-the-shelf architectures. We optimize the SSGNet without any supervision by proposing two novel training losses that enforce task-specific scene structure generation during training.
Our main contribution is to show that such a simple network can achieve state-of-the-art results for several low-level vision applications.
We also demonstrate that our network generalizes well on unseen datasets, compared to existing methods which use structural embedding frameworks. We further propose a lighter version of SSGNet ($\sim$ 29K parameters) for depth computation, SSGNet-D, and successfully execute it on edge computing devices like Jetson AGX Orin, improving the performance of baseline network, even in the wild, with little computational delay.
Our source codes are available at \url{https://github.com/jsshin98/SSGNet}.
\end{abstract}

\begin{graphicalabstract}
\includegraphics[clip=true, width=\linewidth]{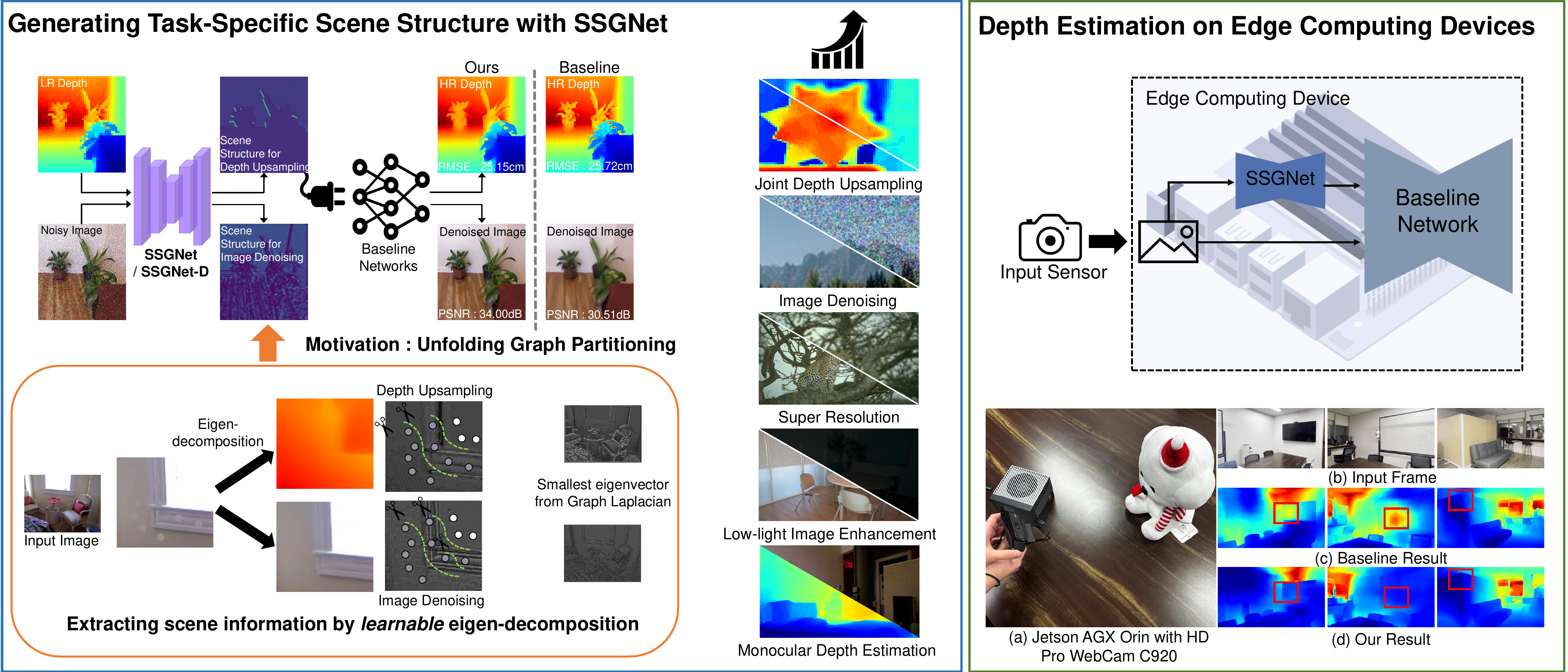}
\end{graphicalabstract}

\begin{highlights}
\item Unfolding graph partitioning: Novel approach for pixel-wise feature learning to extract structural information
\item Lightweight and efficient: Scene Structure Guidance Network (SSGNet) as a plug-and-play module for several low-level vision tasks
\item Unsupervised learning: SSGNet accounts for task-specific scene structures without any supervision
\item Generalizable results: SSGNet demonstrates state-of-the-art performances across a variety of low-level vision applications
\item Edge computing implementation: Low-level vision algorithms with our SSGNet can be embedded into edge computing devices with minimal computational delay

\end{highlights}


\end{frontmatter}


\section{Introduction}
\label{sec:intro}
Methods for estimating scene structures have attracted wide research attention for the past several decades. As an example, texture representations based on image edges have been extensively studied with impressive performance on low-level vision tasks, \ie~image denoising~\citep{tomasi1998bilateral}, deblurring~\citep{krishnan2009fast,levin2007image}, super-resolution~\citep{tai2010super} and inpainting~\citep{nazeri2019edgeconnect,yang2020learning,guo2021image}. Another aspect of scene structures involves inferring robust object boundaries to quantify uncertainty and refine initial predictions in visual perception tasks including joint filtering~\citep{he2012guided,guo2018mutually,li2016deep} and depth completion~\citep{eldesokey2020uncertainty}. Clearly, the goodness of scene structures depends on the target applications, and is defined by either training data or objective functions.
\newline
More recent approaches of extracting informative scene structures have focused on capturing task-specific features from various learning frameworks. One interesting work for joint filtering in \citep{de2022learning} builds graph nodes on learned features from a guidance image to encode semantic information, and represents scene structures by segmenting the graph edges based on objective functions. However, they have heavy computational burdens and are not implemented as an end-to-end architecture.
To formulate an end-to-end architecture, edge priors, directly obtained from conventional edge detection~\citep{irwin1968isotropic,canny1986computational}, are used as a guide. Typically, image edges (or gradients) represent high frequency features and can be forced to generate fine details in the prediction results~\citep{fang2020soft}.
Nevertheless, the question of how to effectively exploit structure guidance information remains unanswered. Tremendous efforts have been made to only generate a single purpose scene structure with each different architecture.

\begin{figure}[t]
    \centering
    \includegraphics[trim={0 4cm 0 0.5cm}, clip=true, width=\linewidth]{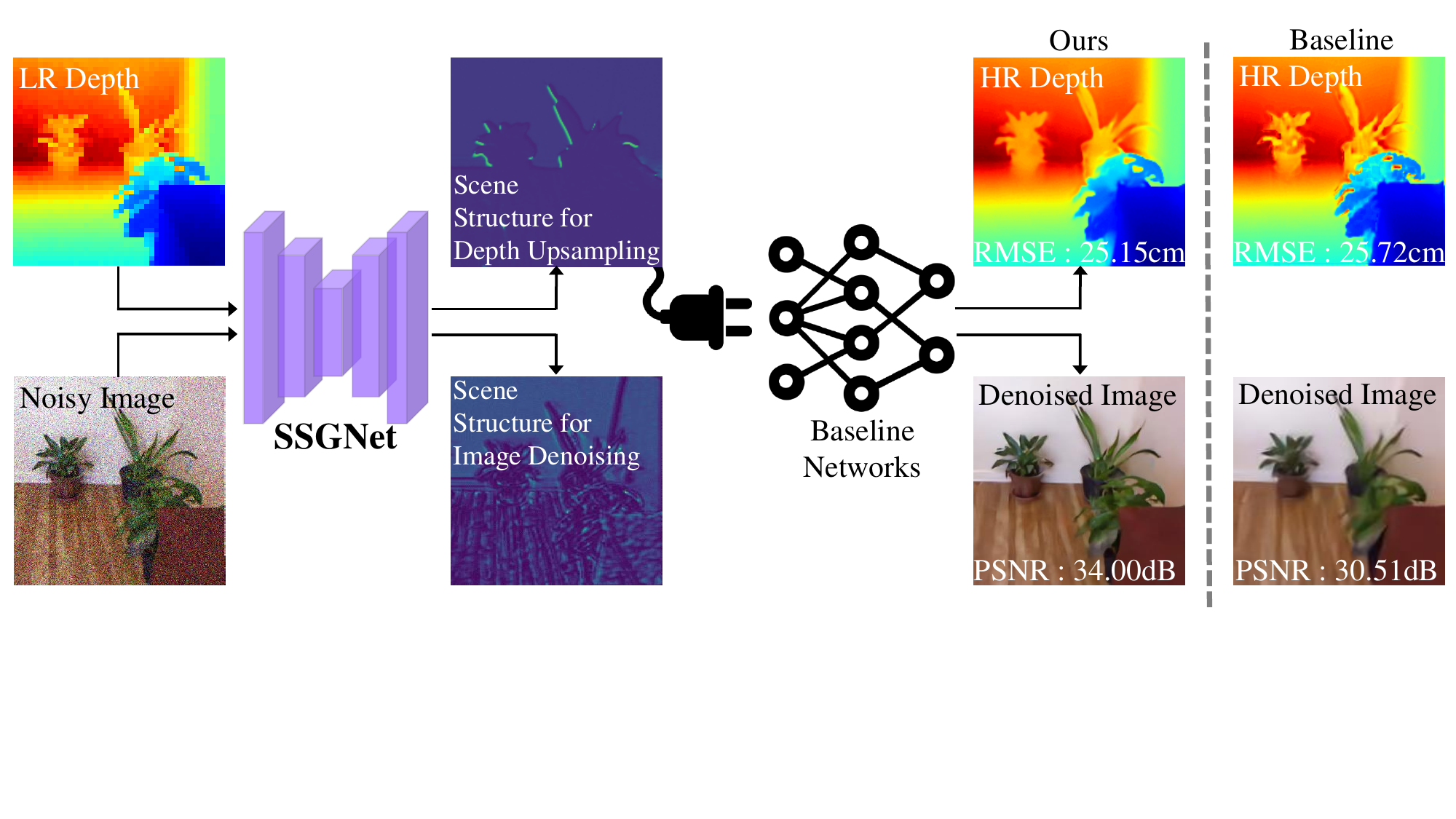}
    \vspace{-7mm}
    \caption{Our SSGNet is a lightweight architecture and can be applied as a plug-and-play module to improve the performance of baseline networks for low-level vision tasks.}
    \label{teaser}
\end{figure}

In this paper, we propose a \textit{Scene Structure Guidance Network (SSGNet)}, a single general neural network architecture for extracting task-specific structural features of scenes. Our SSGNet is lightweight in both size and computation, and is a plug-and-play module that can be applied to any baseline low-level vision architectures.
The SSGNet computes a set of parameterized eigenvector maps, whose combination is selectively determined in favor of the target domain. To achieve this, we introduce two effective losses:
(1) \textit{Eigen loss}, motivated by the traditional graph partitioning problem~\citep{shi2000normalized}, forms a basis set of scene structures based on weight graphs on an image grid. (2) \textit{Spatial loss} enforces the sparsity of each eigenvector for diverse representations of scene structures. We note that, without any supervision, our SSGNet can successfully learn to generate task-specific and informative structural information as shown in \cref{teaser}.

To demonstrate the wide applicability of our SSGNet, we conduct extensive experiments on several low-level vision applications and achieve state-of-the-art results, even in zero-shot cross validations. 
We further embed our structural guidance module on edge computing devices. To do this, we additionally design a lighter-weight module for depth computations, SSGNet-D, in that those tasks require relatively sparse information such as object boundary, compared to image restoration tasks where estimate whole image edges. 
With only 29K parameters, SSGNet-D effectively extracts the object boundaries of scenes and brings the performance gain on depth computations. We implement our newly proposed SSGNet-D on the edge computing device (Jetson AGX Orin) and check the performance gain on monocular depth estimation task in the wild, with little computational delay.

\section{Related work}
Our work is closely related to scene structure embedding for low-level vision tasks.

\subsection{Low-level vision tasks}
The goal of low-level vision tasks such as denoising or super-resolution is to recover a sharp latent image from an input image that has been degraded by the inherent limitations of the acquisition systems (\ie~sensor size, depth of field or light efficiency). In the past decade, there have been significant improvements in low-level vision tasks, and recently deep learning-based techniques have especially proven to be powerful systems.

With the help of inductive bias~\citep{cohen2016inductive}, convolutional neural networks (CNNs) with a pixel-wise photo consistency loss~\citep{li2016deep, zhang2018generalized, zhong2021pixel} are adopted. 
To mitigate the issue on inter-pixel consistency on CNNs, generative adversarial networks (GANs)~\citep{goodfellow2014generative, zhu2017unpaired, karras2019style, liu2021pd, wang2021real}-based methods are proposed to produce visually pleasing results with perceptual losses~\citep{johnson2016perceptual, fuoli2021fourier,  suvorov2022resolution} based on high-level semantic features.
Nowadays, a vision transformer (ViT)~\citep{dosovitskiy2020image, liu2021swin, caron2021emerging, chen2021pre} has been used to capture both local and global image information by leveraging the ability to model long-range context.

Such approaches have shown good progress with structural details. For regularization, adding robust penalties to objective functions~\citep{tibshirani1996regression, xu2010l1,loshchilov2017decoupled, de2022learning} suppresses high-frequency components, and hence the results usually provide a smooth plausible reconstruction.

\subsection{Structural information}
Extensive studies on low-level vision have verified the feasibility and necessity of the image prior including image edges and gradients. 
One of the representative works involves joint image filters which leverage a guidance image as a prior and transfer its structural details to a target image for edge-preserved smoothing~\citep{tomasi1998bilateral,he2012guided,zhang2014rolling}.

Such structure information can be defined in practice, depending on the tasks.
Both super-resolution \citep{pickup2003sampled,sun2008image,xie2015edge,fang2020soft} and image denoising \citep{liu2020gradnet}, which utilize a patch similarity, generate gradient maps to reconstruct high frequency details or suppress image noises.
Works in~\citep{gu2017learning,jin2020geometric} infer object boundaries to refine initial predictions in visual perception tasks, including depth estimation/completion. Also, image inpainting \citep{nazeri2019edgeconnect,yang2020learning,guo2021image,cao2021learning}, filling in missing parts of corrupted scenes, adopt edge maps from traditional method like Canny edge detector~\citep{canny1986computational} to hallucinate their own scene structures.

In spite of promising results from the state-of-the-art methods learning meaningful details for each task, they require a high modeling capacity with numerous parameters and ground-truth structure maps for training. In contrast, our SSGNet, a very small network generating scene structures without any supervision, has advantages for various low-level vision tasks, simply by embedding as an additional module.

\section{Methodology}
Motivated by spectral graph theory~\citep{shi2000normalized, levin2008spectral, levin2007closed}, a set of basis represents scene configurations as a linear combination of the basis. Such parameterization provides a restrictive solution space to accommodate semantic entities like textures and object boundaries. 
Following the works in~\citep{tang2018ba, bloesch2018codeslam}, we begin with an introduction to spectral methods, and then parameterize scene structures which can be used as guidance for various vision tasks. 

\subsection{Motivation}
\label{sec:motivation}
Let us set a weighted undirected graph $\textbf{G} = (\textbf{V}, \textbf{E})$ in an arbitrary feature space with a set of nodes $\textbf{V}$, and a set of edges $\textbf{E}$, whose weight can be represented as an $N \times N$ non-negative adjacency matrix $\textbf{W} = \{w(i, j) : (i, j) \in \textbf{E}\}$ where $i,j$ denote graph nodes.
The Laplacian matrix $L$ of this graph is then obtained by $\textbf{L} = \textbf{D} - \textbf{W}$, where $\textbf{D}$ is a diagonal matrix with the row-wise sum of $\textbf{W}$ on its diagonal. Since the Laplacian matrix is a positive semidefinite matrix, for every $N$ dimensional vector $y$ from the matrix $\textbf{Y}$ which consists of a set of vectors, it holds that 
\begin{equation} 
\label{xlx}
    y^T\textbf{L}y = \sum_{(i, j) \in \textbf{E}}{w(i, j)\{y(i)-y(j)}\}^2 \ge 0.
\end{equation}
To minimize the \cref{xlx}, the indicator vector $y$ should take similar values for nodes $i$ and $j$. When the adjacent value $w(i, j)$ is high, the two nodes are more tightly coupled.

Spectral graph theory in \citep{fiedler1973algebraic, shi2000normalized} proves that the eigenvectors of the graph Laplacian yield minimum-energy graph partitions, and each smallest eigenvector, like the indicator vector $y$, partitions the graph into soft-segments based on its adjacent matrix.

In the image domain, a reference pixel and its similarity to neighboring pixels can be interpreted as a node and edges in a graph~\citep{boykov2001fast}, respectively. In general, affinity is defined by appearance similarities (\ie ~the absolute of intensity differences). With this motivation, images can be decomposed into soft image clusters from a pre-computed affinity matrix. In addition, scene configurations in images can be described as a set of eigenvectors whose smallest eigenvalues indicate connected components on the affinity matrix.

\begin{figure*}[t]
    \centering
    \includegraphics[clip, width=\linewidth]{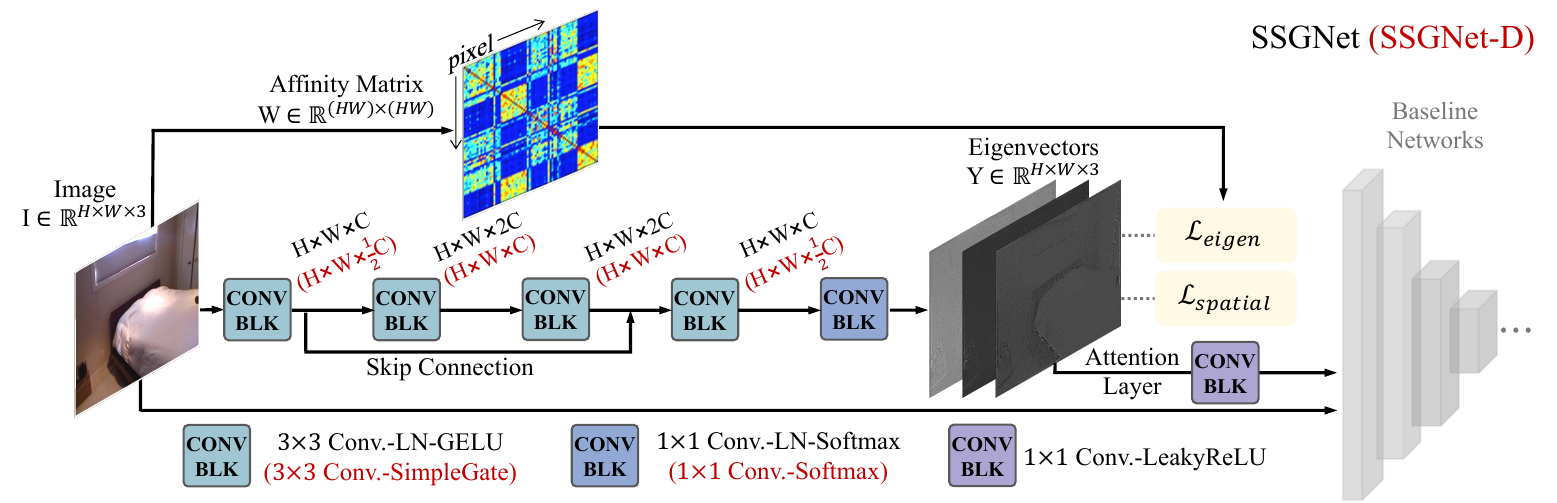}
    \vspace{-5mm}
    \caption{An overview of SSGNet. LN, GeLU, and LeakyReLU denote the layer normalization, GeLU activation, and LeakyReLU activation, respectively. The eigenvectors are integrated via the attention layer, and then embedded to any baseline network. Note that SSGNet-D will be further described in \cref{sec:jetson}.} 
    \label{ssgnet}
\end{figure*}

\subsection{Scene Structure Guidance Network}
In this work, our goal is to train the proposed network, SSGNet, without any supervision because it is infeasible to define a unique objective function for a task-specific structure guidance. 
To accomplish this, we devise a learnable and parametric way of efficiently representing scene structures.

Given single color images $\textbf{I}$~$\in$~$\mathbb{R}^{h\times w\times 3}$, our SSGNet $\sigma$ yields a set of eigenvectors $\textbf{Y}\in\mathbb{R}^{h\times w\times n}$, where $n$ denotes the number of eigenvectors and is empirically set to 3:
\begin{equation}
    \textbf{Y} = \sigma(\textbf{I}).
\end{equation}
As illustrated in \cref{ssgnet}, SSGNet takes a simple encoder-decoder architecture ($\sim$ 56K), consisting of two 3$\times$3 convolutional layers and three 3$\times$3 deconvolutional layers with layer normalizations~\citep{ba2016layer} and gelu activations~\citep{hendrycks2016gaussian} after each layer except for the last softmax layer. 
The output of our SSGNet is associated with learnable weights that will be finetuned in accordance with an objective function of target applications.

To optimize SSGNet in an unsupervised manner, we define a loss function $\mathcal{L}_{ssg}$ which is a linear combination of two loss terms as follows:

\subsubsection{Eigen Loss}
The main objective of SSGNet is to obtain a set of smallest eigenvectors $\textbf{Y}$ of the graph Laplacian $\textbf{L}$, inspired by the spectral graph theory~\citep{fiedler1973algebraic,shi2000normalized,levin2008spectral}.

To generate the graph Laplacian $\textbf{L}$, we trace back all the way down to some traditional similarity matrix methods.
Since an image is segmented based on a constructed affinity matrix in spectral graph theory, the form of the matrix depends on the pixel-level similarity encoding~\citep{levin2008spectral,levin2007closed,chen2013knn}. 
In this work, we adopt the sparse KNN-matting matrix~\citep{chen2013knn}. To be specific, we first collect nonlocal neighborhoods $j$ of a pixel $i$ by the k-nearest neighbor algorithm (KNN)~\citep{cover1967nearest}. Then, we define the feature vector $\varphi(i)$ at a given pixel $i$ as follows:
\begin{equation}
\label{knn}
    \varphi(i) = (r, g, b, d_x, d_y)_i,    
\end{equation}
where $(r, g, b)$ denotes each color channel, and $(d_x, d_y)$ is a weighted spatial coordinate for the $x$- and $y$-axes.
We follow the KNN kernel function $\text{KNN}(i)$ to construct the sparse affinity matrix $\textbf{W}$ based on feature vectors $\varphi$:
\begin{eqnarray}
    \textbf{W}(i, j)=\left\{
    \begin{array}
        {ll}1 - \parallel \varphi(i) - \varphi(j) \parallel , & j \in \text{KNN}(i)
            \\ 0, & \text{otherwise},
    \end{array}\right.
\end{eqnarray}
where $j \in \text{KNN}(i)$ are the k-nearest neighbors of $i$ based on the distance defined by $\varphi$. Using the sparse KNN-matting matrix, we can take account of both spatial distance and color information with less computational cost than a traditional similarity matrix.
The graph Laplacian $\textbf{L}$ is finally obtained by $\textbf{L} = \textbf{D} - \textbf{W}$ as the same manner, described in~\cref{sec:motivation}.

We can finally obtain a set of eigenvectors $\textbf{Y}$ by minimizing the quadratic form of $\textbf{L}$, $\mathcal{L}_{eigen}$, as below:
\begin{equation}
\label{eqn1}
    \mathcal{L}_{eigen} = \sum_{k}\textbf{Y}_k^T\textbf{L}\textbf{Y}_k.
\end{equation}

However, we observe that SSGNet sometimes produces undesirable results during the training phase because of the degenerate case, where all eigenvectors have the same value, and needs an additional loss term to regularize it.

\subsubsection{Spatial Loss}
Since our SSGNet uses a softmax function in the last layer to prevent the eigenvectors from converging to zero vectors, we only need to handle the degenerate case.
Our spatial loss $\mathcal{L}_{spatial}$ considers the sparsity of each eigenvector to enforce diverse representations of scene structure, defined as below:
\begin{equation}
\label{eqn2}
    \mathcal{L}_{spatial} = \sum_k(|\textbf{Y}_k|^\gamma + |1-\textbf{Y}_k|^\gamma)-1,
\end{equation}
where $|\cdot|$ indicates an absolute value, and the hyperparameter $\gamma$ is set to 0.9 in our implementation.
We can intuitively figure out that $\mathcal{L}_{spatial}$ has a minimum value when $\textbf{Y}_k$ is either 0 or 1 for each pixel. With the $\mathcal{L}_{spatial}$ and the softmax operation together, we show that if a pixel of one eigenvector converges near to 1, the pixel of other eigenvectors should go to 0. This makes each pixel across the eigenvectors have different value due to the sparsity penalty, which produces diverse feature representations of image structures.

In total, the final loss function for SSGNet is defined as:
\begin{equation}
    \mathcal{L}_{ssg} =\mathcal{L}_{eigen} + \lambda \mathcal{L}_{spatial}
    \label{eq:loss}
\end{equation}
where $\lambda$ is the hyper-parameter, and is empirically set to  40.
\begin{figure*}[t]
    \centering
    \includegraphics[trim={0 0 0 0}, clip=true, width=\linewidth]{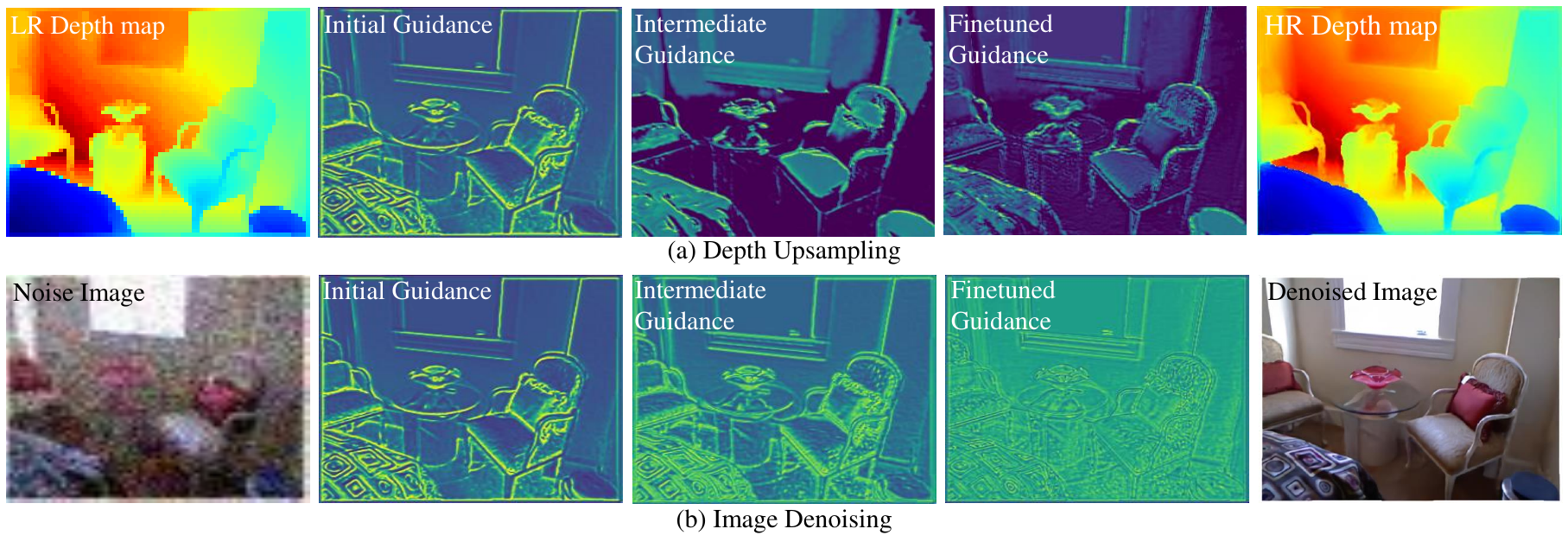}
    \vspace{-7mm}
    \caption{Examples of task-specific scene structures: initial, intermediate and final results from SSGNet for (a) joint depth upsampling and (b) image denoising.}
    \label{iteration}
\end{figure*}

\begin{figure}[t]
    \centering
    \includegraphics[clip=true, width=0.5\linewidth]{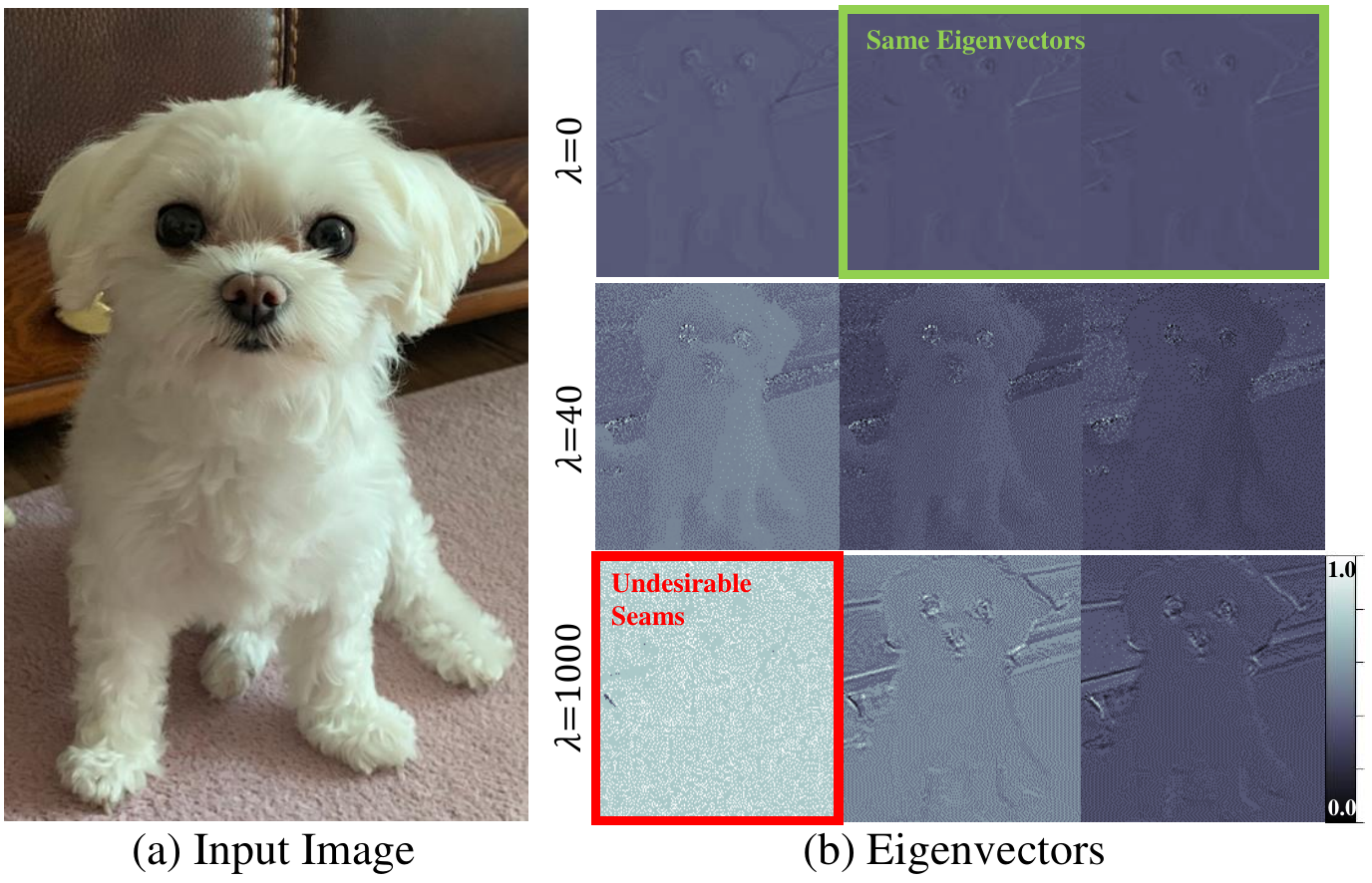}
    \vspace{-5mm}
    \caption{Visualization of the sets of eigenvectors according to $\lambda$ = 0, 40, and 1000.}
    \label{lambda}
\end{figure}

Our SSGNet is first pretrained on a single dataset and can be embedded in various baseline networks and jointly optimized with the baseline after passing through an additional single convolution layer which acts as an attention module. 
In favor of the target domain on each task, this layer produces adaptive structural information of input scenes by linearly combining the set of eigenvectors.

\subsection{Analysis}
To the best of our knowledge, the SSGNet is the first to unfold the eigen-decomposition problem into a learnable network. 
To validate its effectiveness, we provide a series of analyses on SSGNet.

First, we analyze our loss function in \cref{eq:loss} by tuning the hyper-parameter $\lambda$ used as a balancing term between $\mathcal{L}_{spatial}$ and $\mathcal{L}_{eigen}$. In our experiment, the best performance is obtained with $\lambda=40$.
In \cref{lambda}, we show the visualization results for three different $\lambda$ values, including $\lambda$ = 0, 40, and 1000. 
When $\lambda$ is set to 0, $\mathcal{L}_{spatial}$ is not forced enough to give a sparsity penalty across eigenvectors, which leads to the degenerate case. 
Otherwise, if $\lambda$ is set to 1000, the image is not well-segmented because the overwhelming majority of $\mathcal{L}_{spatial}$ causes undesirable seams on the image.

Next, we demonstrate that our SSGNet yields task-specific structural guidance features. As we highlighted, the SSGNet can be embedded in baseline networks. 
When the pretrained SSGNet is attached to baseline networks, the network parameters on SSGNet are finetuned to produce guidance features suitable for each task as the training proceeds. In \cref{iteration}, we visualize how the eigenvectors from SSGNet change at each iteration during finetuning, including joint depth upsampling~\citep{dong2022learning} and single image denoising~\citep{zhang2022idr}.

The joint depth upsampling needs accurate object boundaries as a prior~\citep{li2014similarity}. For obvious reasons, an objective function in the joint depth upsampling encourages a greater focus on reconstructing object boundaries. As shown in \cref{iteration}(a), our SSGNet generates attentive features on them during fine-tuning.
In addition, for image denoising, it is essential to preserve fine detailed textures. In \cref{iteration}(b), with the meaningful scene structures from our SSGNet, the plausible result is inferred as well.
We claim that it is possible for our SSGNet to capture informative and task-specific structures through gradient updates from backpropagation~\citep{lecun1989backpropagation}.
We will describe the experimental details and SSGNet's quantitative benefits on each task in \cref{exp}.

\begin{figure*}[t]
    \centering
    \includegraphics[clip=true, width=\linewidth]{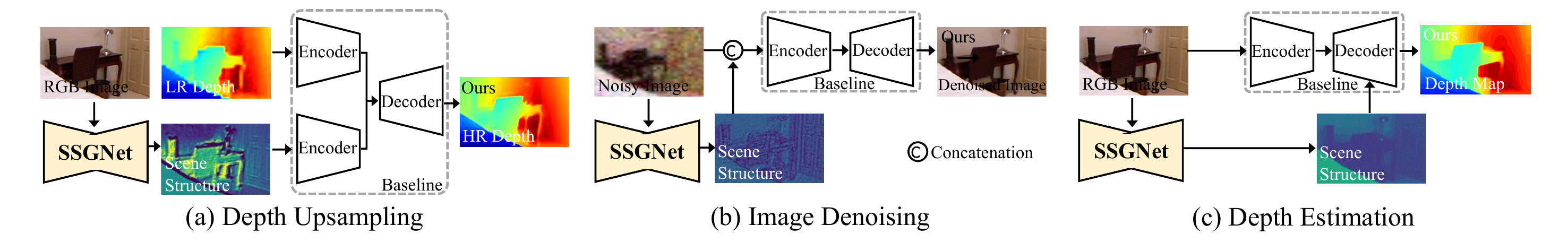}
    \vspace{-7mm}
    \caption{Illustrations of SSGNet for low-level vision tasks. The yellow colored networks indicate our SSGNet that outputs informative task-specific structure guidances.}
    \label{experiment}
\end{figure*}



\subsection{Training Scheme}
We implement the proposed framework using a public Pytorch~\citep{paszke2019pytorch}, and utilize the Adam~\citep{kingma2014adam} optimizer with $\beta_1=0.9$ and, $\beta_2=0.999$. 
The learning rate and the batch size are set to $0.0001$ and 4.
We train the proposed framework on images with a 256$\times$256 resolution. Since the proposed framework consists of fully convolutional layers, images with higher resolutions than that used in the training phase are available in inference.
The training on SSGNet took about 10 hours on two NVIDIA Tesla v100 GPUs.

We conduct a variety of experiments on low-level vision tasks, including self-supervised joint depth upsampling (\cref{depthup_sec}), unsupervised single image denoising (\cref{denoising_sec}), and other various low-level vision tasks (\cref{lowlevel_sec}), including super-resolution, low-level image enhancement and monocular depth estimation, to demonstrate the effectiveness of our SSGNet. 
Moreover, we provide an extensive ablation study (\cref{sec:ablation}) to precisely describe the effects of each component in SSGNet. 
We also show that our lighter-weight module for depth computation, SSGNet-D, further improves the performance of joint depth upsampling (\cref{sec:ssgnetd}) and it can be applied to edge computing devices for image regression tasks like depth estimation (\cref{sec:sub_jetson}) with marginal increase of computational speed.

\subsubsection{Baselines with SSGNet}
In this section, our goal is to validate a wide applicability of SSGNet. To do this, we incorporate SSGNet into existing CNN architectures for the joint depth upsampling and the unsupervised image denoising by simply embedding scene structures from ours to the models. As shown in \cref{experiment}, we incorporate our SSGNet into a variety of baseline networks at diverse locations, demonstrating its adaptability and efficiency within different network structures, while preserving the fundamental pipeline of the baseline systems.

Prior to the evaluations, we train our SSGNet on a well-known NYUv2 dataset \citep{silberman2011indoor}, consisting of 1,000 training images and 449 test images. With the pre-trained weight of SSGNet, we embed it to the baseline networks and finetune on each task. As mentioned above, we do not need any supervision for training SSGNet. 
We note that NYUv2 dataset is not used for evaluations, to validate the zero-shot generalization across various datasets. 



\section{Experiments}
\label{exp}

\subsection{Joint Depth Upsampling}
\label{depthup_sec}

Joint depth upsampling leverages the explicit structure detail of the input image as a guidance and transfers it to the target low-resolution depth map for enhancing spatial resolution. With this application, we demonstrate the synergy of the structure details from clean input images and the proposed learnable scene structure from SSGNet.

For this experiment, we choose MMSR~\citep{dong2022learning} as a baseline depth upsampling network. MMSR introduces a mutual modulation strategy with the cross-domain adaptive filtering and adopts a cycle consistency loss to train the model in a fully self-supervised manner. Instead of directly using the input image as the guidance, we employ the structure guidance from the pretrained SSGNet in~\cref{experiment}(a), and follow the training scheme of MMSR for fair comparisons such that all the supervised methods are trained on NYUv2 dataset. Note that the results from SSGNet equipped are denoted as Ours.

We also follow the evaluation protocol described in~\citep{dong2022learning} to quantitatively measure the root mean square error (RMSE) and the mean absolute error (MAE). To be specific, we use the Middlebury stereo dataset 2005~\citep{scharstein2007learning}, 2006~\citep{hirschmuller2007evaluation}, and 2014~\citep{scharstein2014high}\footnote{Since Middlebury 2003 provides neither depth maps nor camera parameters, we could not use it in this evaluation.}, and augment them, which provides 40, 72, and 308 image-depth pairs, respectively, using a public code\footnote{Downloaded from \url{https://rb.gy/bxyqgi}}.

\begin{figure*}[t]
    \centering
    \includegraphics[clip=true, width=\linewidth]{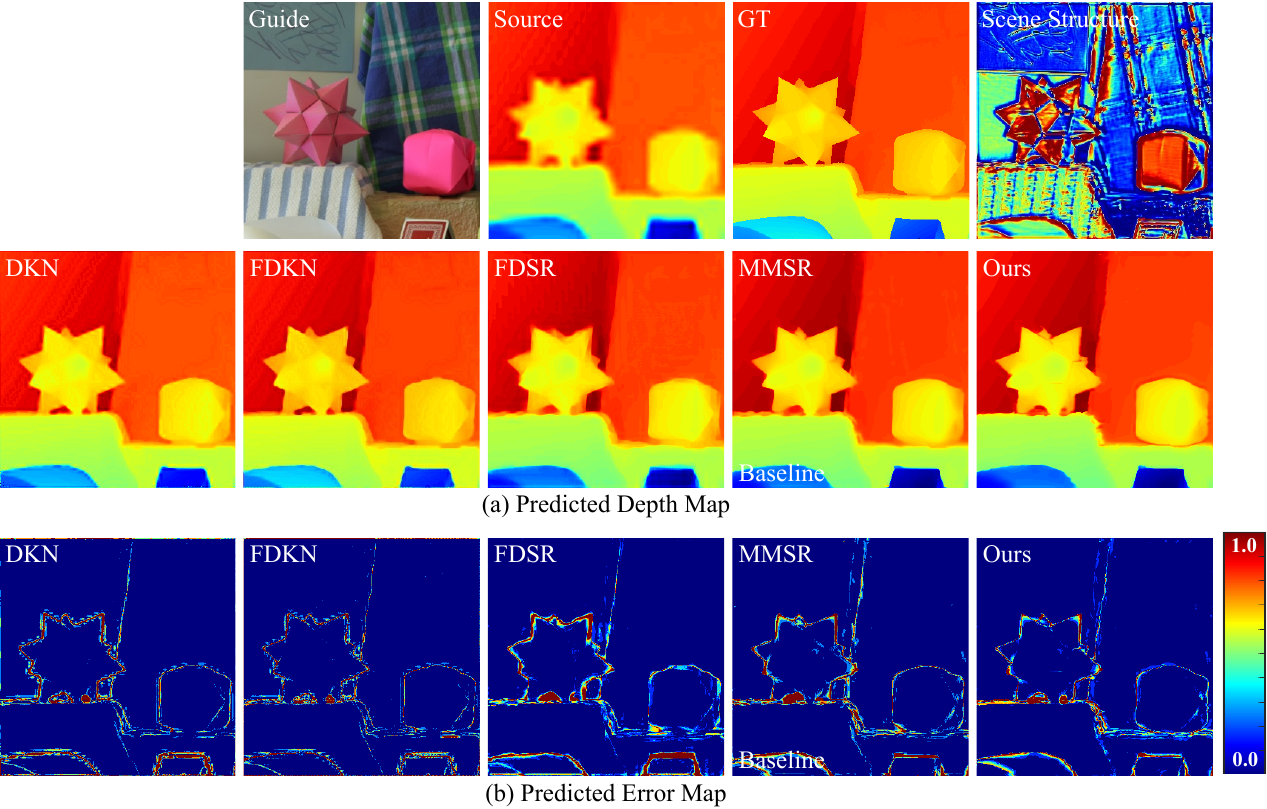}
    \vspace{-5mm}
    \caption{Comparison results on the joint depth upsampling with a resolution factor of 8 on the Middlebury 2005 dataset. We visualize the predictions and their corresponding error maps of competitive methods and ours. The results from SSGNet equipped are denoted as Ours.}
    \label{depth_figure}
\end{figure*}

\begin{table*}[h]
\caption{Quantitative results on joint depth upsampling tasks. The best and the second best results are marked as \textbf{bold} and \underline{underlined}, respectively. (unit:cm) The results from SSGNet are denoted as Ours.}
\centering
\tiny
\resizebox{\linewidth}{!}{%
\begin{tabular}{ccc|ccc|ccc}
\hline\hline
\multicolumn{2}{c}{} & \multicolumn{1}{c|}{} & \multicolumn{3}{c|}{Supervised} & \multicolumn{3}{c}{Self-Supervised} \\
\hhline{|~~~|------|}
Dataset & Scale & \multicolumn{1}{c|}{} & DKN & FDKN & \multicolumn{1}{c|}{FDSR} & P2P & MMSR & Ours\\ 
\hline
\multirow{2}{*}{2005} & \multirow{2}{*}{$\times$4} & RMSE & 1.103 & 0.964 & 0.886 & 1.288 & \underline{0.708} & \textbf{0.612}\\
                      & & MAE & 0.275 & 0.222 & \underline{0.211} & 0.273 & 0.239 & \textbf{0.188}\\ 
\hhline{|~~|-------|}
& \multirow{2}{*}{$\times$8} & RMSE & 1.182 & 1.629 & \underline{1.043} & 1.177 & \underline{1.043} & \textbf{0.830}\\
                      & & MAE & 0.288     & 0.339    & 0.333    & \underline{0.280}      & 0.319      & \textbf{0.245}\\ 
\hline
\multirow{2}{*}{2006} & \multirow{2}{*}{$\times$4} & RMSE &                            1.623 & 1.337 & 1.198 & 2.604 & \underline{0.555} & \textbf{0.504}\\
                      & & MAE & 0.297 & 0.222 & \textbf{0.198} & 0.413 & 0.232 & \underline{0.201}\\
\hhline{|~~|-------|}
                      & \multirow{2}{*}{$\times$8} & RMSE & 1.790 & 1.883 & 1.170 & 2.684 & \underline{0.723} & \textbf{0.648}\\
                      & & MAE & 0.307 & 0.305 & 0.267 & 0.300 & \underline{0.261} & \textbf{0.225}\\ 
\hline
\multirow{2}{*}{2014} & \multirow{2}{*}{$\times$4} & RMSE & 2.878 & 2.593 & 3.217 & 4.019 & \underline{1.953} & \textbf{1.819}\\
                      & & MAE & 0.739 & 0.659 & 0.595 & 0.822 & \underline{0.573} & \textbf{0.451}\\ 
\hhline{|~~|-------|}
                      & \multirow{2}{*}{$\times$8} & RMSE & 3.642 & 3.510 & 3.606 & 3.894 & \underline{2.765} & \textbf{2.714}\\
                      & & MAE & \underline{0.775} & 0.871 & 0.885 & 0.920 & 0.785 & \textbf{0.675}\\  
\hline\hline
\end{tabular}%
}
\label{depth_table}
\end{table*}

We compare with various state-of-the-art models, including supervised, DKN~\citep{kim2021deformable}, FDKN~\citep{kim2021deformable} and FDSR~\citep{he2021towards}, and self-supervised manners, P2P~\citep{lutio2019guided} and MMSR~\citep{dong2022learning}.
As shown in \cref{depth_table}, MMSR with our SSGNet embedded achieves the best performance in almost datasets over the comparison methods. Our SSGNet brings the performance gain over the second best method (MMSR) is about 10.4$\%$ and 11.8$\%$ with respect to RMSE and MAE, respectively.  It is also noticeable that the scene structure contributes to reducing the errors in the star-like object boundary and the inside surface, visualized in \cref{depth_figure}. We highlight that the result demonstrates the strong generalization capabilities of our SSGNet on unseen data again.



\subsection{Image Denoising}
\label{denoising_sec}
We treat single image denoising to check the effectiveness of our SSGNet if the scene structure in the input image is corrupted by noise.  
For this experiment, we use IDR~\citep{zhang2022idr} as a baseline image denoising network. 
IDR suppresses the image noise in a self-supervised manner by proposing an iterative data refinement scheme. The key of IDR is to reduce a data bias between synthetic-real noisy images and ideal noisy-clean images.
To embed the scene structure to IDR, we simply concatenate it from our pretrained SSGNet with the noisy input image in~\cref{experiment}(b). As the rounds go on iteratively, our SSGNet focuses more on texture information of input scenes by ignoring the image noise, as already displayed in \cref{iteration}.

\begin{figure*}[t]
    \centering
    \includegraphics[clip=true, width=\linewidth]{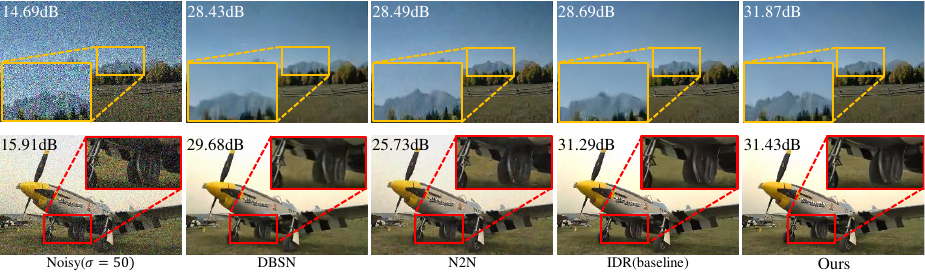}
    \vspace{-7mm}
    \caption{Examples of the single image denoising. For the noisy level $\sigma=50$, we visualize the results from IDR+SSGNet as well as the state-of-the-art methods.}
    \label{denoising_figure}
\end{figure*}

\begin{table*}[t]
\caption{Quantitative results on single image denoising.}
\centering
\normalsize
\setlength\extrarowheight{0.5pt}
\resizebox{\linewidth}{!}{%
\begin{tabular}{c|cccc|cccc|cccc}
\hline
\hline
\multicolumn{1}{c|}{} & \multicolumn{4}{c|}{Kodak}&\multicolumn{4}{c|}{BSD300}&\multicolumn{4}{c}{BSD68}\\
\hhline{|~|------------|}
\multicolumn{1}{c|}{Method} &\multicolumn{2}{c}{$\sigma=25$}&\multicolumn{2}{c|}{$\sigma=50$}&\multicolumn{2}{c}{$\sigma=25$}&\multicolumn{2}{c|}{$\sigma=50$}&\multicolumn{2}{c}{$\sigma=25$}&\multicolumn{2}{c}{$\sigma=50$}\\
\hhline{|~|------------|}
 \multicolumn{1}{c|}{} & PSNR & SSIM & PSNR & \multicolumn{1}{c|}{SSIM}& PSNR & SSIM& PSNR & \multicolumn{1}{c|}{SSIM} & PSNR & SSIM& PSNR & \multicolumn{1}{c}{SSIM}\\
\hline
BM3D  & 31.88 & 0.869& 28.64& 0.772 & 30.47& 0.863 & 27.14 & 0.745 & 28.55& 0.782 & 25.59  & 0.670\\

N2V   & 31.63 & 0.869& 28.57 & 0.776 & 30.72 & 0.874 & 27.60& 0.775 & 27.64 & 0.781 & 25.46& 0.681 \\

Nr2n  & 31.96 & 0.869 & 28.73& 0.770 & 29.57& 0.815 & 26.18& 0.684 & N/A& N/A  & N/A & N/A  \\

DBSN  & 32.07& 0.875 & 28.81& 0.783 & 31.12& 0.881 & 27.87& 0.782 & 28.81& 0.818 & 25.95& 0.703 \\
          
N2N   & \textbf{32.39}& \textbf{0.886} & 29.23& \underline{0.803} & 31.39 & 0.889 & 28.17& 0.799 & 29.15& \underline{0.831} & 26.23& 0.725 \\
\hline
IDR   & \underline{32.36}& 0.884 & \underline{29.27} & \underline{0.803} & \underline{31.48}& \underline{0.890} & \underline{28.25} & \underline{0.802}  & \underline{29.20}& \textbf{0.835} & \underline{26.25}& \underline{0.726}\\
          
Ours   & \textbf{32.39}& \underline{0.885} & \textbf{29.34}& \textbf{0.806} & \textbf{31.52}& \textbf{0.891} & \textbf{28.33}& \textbf{0.805}  & \textbf{29.25} & \textbf{0.835} & \textbf{26.36} & \textbf{0.731}\\
\hline 
\hline
\end{tabular}%
}
\label{denoising_table}
\end{table*}

To validate the applicability to the image denoising task as well, we compare our results with various state-of-the-art self-supervised models, including BM3D~\citep{makinen2019exact} N2V~\citep{krull2019noise2void}, Nr2n~\citep{moran2020noisier2noise} DBSN~\citep{wu2020unpaired}, N2N~\citep{lehtinen2018noise2noise}, and IDR~\citep{zhang2022idr}.
For the evaluation, we strictly follow the experimental setup in~\citep{zhang2022idr}. We quantitatively measure PSNR and SSIM on Kodak~\citep{kodak1993kodak}, BSD300~\citep{movahedi2010design} and BSD68~\citep{martin2001database} datasets for the zero-shot generalization. The models are trained on Gaussian noise with the continuous noise level $\sigma=[0, 50]$ and tested on $\sigma=25$ and 50.

As shown in \cref{denoising_table}, IDR with our SSGNet embedded achieves the best performance among all the competitive methods regardless of the noise levels.
We emphasize that the performance gain by our SSGNet is about 0.58dB on average. Considering the performance difference between the second and the third best methods is about 0.26dB achieved by the paradigm shift from a statistical reasoning of image restoration~\citep{lehtinen2018noise2noise} to the iterative refinement~\citep{zhang2022idr}, SSGNet makes meaningful contribution. \Cref{denoising_figure} shows some example results. With the powerful capability of IDR on the noise suppression, our SSGNet preserves the scene texture of the objects well.

\subsection{Other low-level vision tasks}
\label{lowlevel_sec}
We conduct additional experiments for other low-level vision tasks, including image super-resolution and low-light image enhancement. 
The single image super resolution infers high-resolution images from low-resolution images, and low-light image enhancement aims to improve the visibility of input images captured under low-light conditions, and to suppress image noise and artifacts.
Since structure guidances for image super-resolution and low-light image enhancement tasks are extracted from degraded images, we can further validate the effectiveness of our SSGNet.

For super resolution, we choose FDIWN~\citep{gao2022feature}, a lightweight CNN architecture equipped with feature distillation, for our baseline model. We concatenate the extracted structure guidance from our pretrained SSGNet with RGB image as an input to the network.


In addition, the baseline model for low-light image enhancement is LLFlow~\citep{wang2022low}, consisting of a conditional encoder to extract the illumination-invariant color map and an invertible network that learns a distribution of normally exposed images conditioned on a low-light one. We concatenate the extracted structure guidance from our pretrained SSGNet with the network inputs such as low-light image, color and noise map. To perform zero-shot validation, we use LOL dataset~\citep{wei2018deep} for training and evaluate the model with VE-LOL dataset~\citep{liu2021benchmarking}, following the cross-dataset evaluation scheme of LLFlow~\citep{wang2022low}.

\begin{figure*}[t]
    \centering
    \includegraphics[clip=true, width=\linewidth]{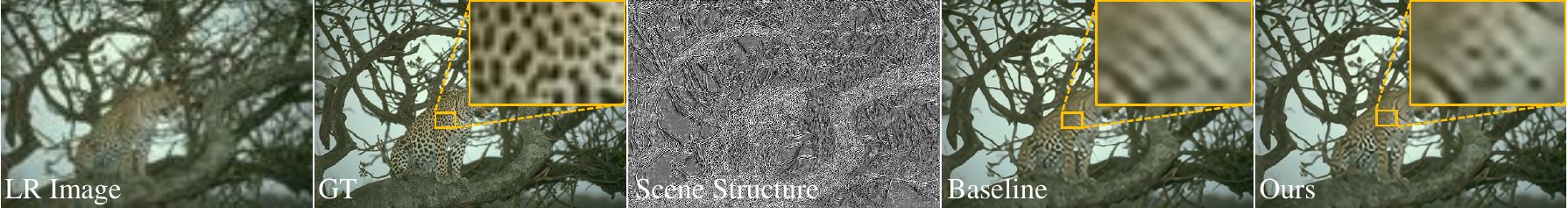}
    \caption{Comparison results on single image super-resolution on the B100 dataset with FDIWN.}
    \label{fdiwn}
\end{figure*}

\begin{figure*}[t]
    \centering
    \includegraphics[clip=true, width=\linewidth]{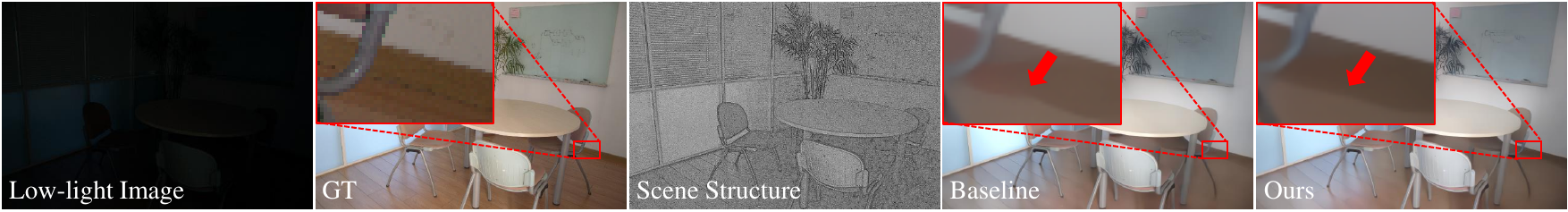}
    \caption{Comparison results on low-light image enhancement on the VE-LOL dataset with LLFlow.}
    \label{llflow}
\end{figure*}

Note that we use the author-provided official codes and train both baseline and ours from scratch for a fair comparison.
\Cref{fig:lowlevels} reports that we obtain the performance gains with our SSGNet over the baseline models. \Cref{fdiwn}, \cref{llflow} show some example results for image super-resolution and low-light image enhancement. The synergy between the baseline models and our SSGNet is noticeable. 

\begin{figure}[t]
    \centering
    \includegraphics[clip=true, width=\linewidth]{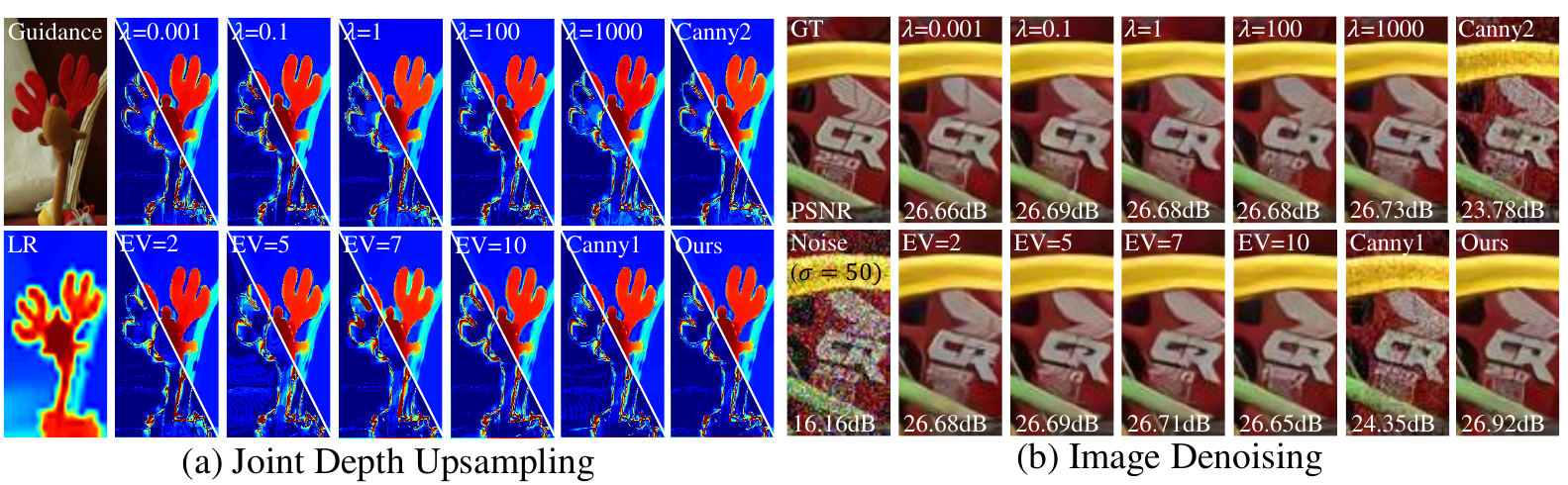}
    \caption{Qualitative comparison for different settings of SSGNet. EV denotes the number of eigenvectors, and Canny1 and Canny2 mean edge threshold settings such as $\psi$ = \{0.6, 0.9, 1.4\} and $\psi$ = \{1.0, 2.0, 3.0\}, respectively. For the joint depth upsampling, we display the reconstruction results and the error maps, together.}
    \label{denoising_ablation}
\end{figure}

\newcommand{\mcx}[3]{\multicolumn{#1}{#2c}{#3}}
\newcommand{\spx}{\kern2ex}

\begin{table}[t]
\caption{Additional experiments on Set5~\citep{bevilacqua2012low}, Set14~\citep{zeyde2012single} and Manga109~\citep{matsui2017sketch} for image super-resolution with $\times$2 and VE-LOL~\citep{liu2021benchmarking} dataset for low-light image enhancement.}
\centering
\tiny
\resizebox{0.9\linewidth}{!}{%
\begin{tabular}{ccccccccccccc}
\hline
\multirow{2}{*}{Methods} & \mcx{4}{|}{Set5} & \mcx{4}{|}{Set14} & \mcx{4}{|}{Manga109} \\
\hhline{|~~|-----------|}
&  \mcx{2}{|}{PSNR} & \mcx{2}{}{SSIM} & \mcx{2}{|}{PSNR} & \mcx{2}{}{SSIM} & \mcx{2}{|}{PSNR} & \mcx{2}{}{SSIM}\\
\hline
FDIWN & \mcx{2}{|}{32.23} & \mcx{2}{}{0.896} & \mcx{2}{|}{28.66} & \mcx{2}{}{0.738} & \mcx{2}{|}{30.63} & \mcx{2}{}{0.910}\\
Ours & \mcx{2}{|}{\textbf{32.27}} & \mcx{2}{}{\textbf{0.904}} & \mcx{2}{|}{\textbf{28.68}} & \mcx{2}{}{\textbf{0.757}} & \mcx{2}{|}{\textbf{30.66}} & \mcx{2}{}{\textbf{0.914}}\\
\hline    
\hline
\multirow{2}{*}{Methods} & \mcx{12}{|}{VE-LOL (Zero-shot validation)} \\
\hhline{|~~|-----------|}
                        & \mcx{4}{|}{PSNR} & \mcx{4}{|}{SSIM} & \mcx{4}{|}{LPIPS}\\
\hline
LLFlow & \mcx{4}{|}{22.38} & \mcx{4}{|}{0.7416} & \mcx{4}{|}{0.3194} \\
Ours & \mcx{4}{|}{\textbf{22.51}} & \mcx{4}{|}{\textbf{0.7682}} & \mcx{4}{|}{\textbf{0.2778}} \\
\hline
\end{tabular}
}
\label{fig:lowlevels}
\end{table}

\subsection{Ablation Study}
\label{sec:ablation}
An extensive ablation study is conducted to examine the effect of each component on SSGNet: the hyper-parameter $\lambda$ in our loss function and the number of eigenvectors. We additionally test alternative scene structures computed from Canny Edge~\citep{canny1986computational} with different thresholds. For this ablation study, we measure RMSE and MAE on the Middlebury 2005 dataset for the joint depth upsampling ($\times8$), and PSNR and SSIM on the Kodak dataset for the single image denoising ($\sigma=50$), whose results and examples are reported in \cref{ablation} and \cref{denoising_ablation}, respectively.


\subsubsection{Choice of Hyper-parameter $\lambda$}
Since our loss function requires the selection of a hyper-parameter $\lambda$, it is important to study the sensitivity of the performances to the choice of $\lambda$. We carry out this experiment for six different values: 0.001, 0.1, 1, 100 and 1000 as well as 40 in our setting. 

As a result, SSGNet's performance is insensitive to the choice of $\lambda$. In the joint depth upsampling, the performance difference according to $\lambda$ is very marginal in that RMSE and MAE are at most 0.02cm and 0.001cm off the optimal values. In contrast, the performance gain for the image denoising when using $\lambda=40$ is relatively large. Compared to $\lambda=1000$ which shows the second best performance, the improvement from 0.08dB in PSNR when using $\lambda=40$ brings more benefits for the comparisons with the state-of-the-art methods. In total, we find the optimal trade-off between these two tasks.


\subsubsection{The Number of Eigenvectors}
The number of eigenvectors to represent scene structures is closely related to the number of learnable parameters in SSGNet. It is important for us to determine the optimal trade-off parameter in consideration of both the minimum number and the performances on these two tasks.

We investigate the performances of SSGNet with two, five, seven and ten as well as three eigenvectors. Interestingly, we observe the similar phenomenon just as above. The performance degradation on the joint depth upsampling is very small (about 0.02cm in RMSE and 0.003cm in MAE), and the performance gain by 0.07dB in PSNR over the second best value on the image denoising is achieved. For the same reason, we set the number of eigenvectors to 3.


\begin{table*}[h]
\caption{Ablation study for the effects of each component of SSGNet. We use the Middlebury 2005 with $\times$8 for the joint depth upsampling, and the Kodak with a noise level $\sigma$=50 for the single image denoising.}
\centering
\tiny
\resizebox{0.9\linewidth}{!}{%
\begin{tabular}{c|cc|cc}
\hline
\multirow{2}{*}{Task} & \multicolumn{2}{c}{Depth Upsampling}&\multicolumn{2}{c}{Denoising} \\
\hhline{|~|----|}
                      &  ~RMSE~ & ~MAE~ & ~PSNR~ & ~SSIM~ \\
\hhline{|~|----|}
Ours & ~0.83~ & ~0.245~ & ~29.34~ & ~0.806~\\
\hline 
\multicolumn{5}{c}{Hyper-parameter $\lambda$}\\
\hline
0.001 & ~0.84~ & ~0.246~ & ~29.17~ & ~0.801~ \\
0.1 & ~0.83~ & ~0.245~ & ~29.15~ & ~0.800~ \\
1 & ~0.82~ & ~0.244~ & ~29.13~ & ~0.800~ \\
100 & ~0.81~ & ~0.244~ & ~29.16~ & ~0.801~ \\
1000 & ~0.84~ & ~0.248~ & ~29.26~ & ~0.80~ \\
\hline
\multicolumn{5}{c}{$\#$ of eigenvectors}\\
\hline
2 & ~0.84~ & ~0.272~&~29.15~&~0.800~ \\
5 & ~0.81~ & ~0.242~&~29.16~&~0.800~ \\
7 & ~0.82~ & ~0.242~&~29.17~&~0.800~ \\
10 & ~0.82~ & ~0.244~&~29.27~&~0.804~ \\
\hline
\multicolumn{5}{c}{Canny Edge}\\
\hline
$\psi$ = \{0.6, 0.9, 1.4\} & ~0.90~& ~0.298~ &~24.86~ & ~0.510~\\
$\psi$ = \{1.0, 2.0, 3.0\} & ~0.90~& ~0.282~ &~24.37~ & ~0.489~\\
\hline
\end{tabular}%
}
\label{ablation}
\end{table*}

\begin{figure*}[t]
    \centering
    \includegraphics[clip=true, width=\linewidth]{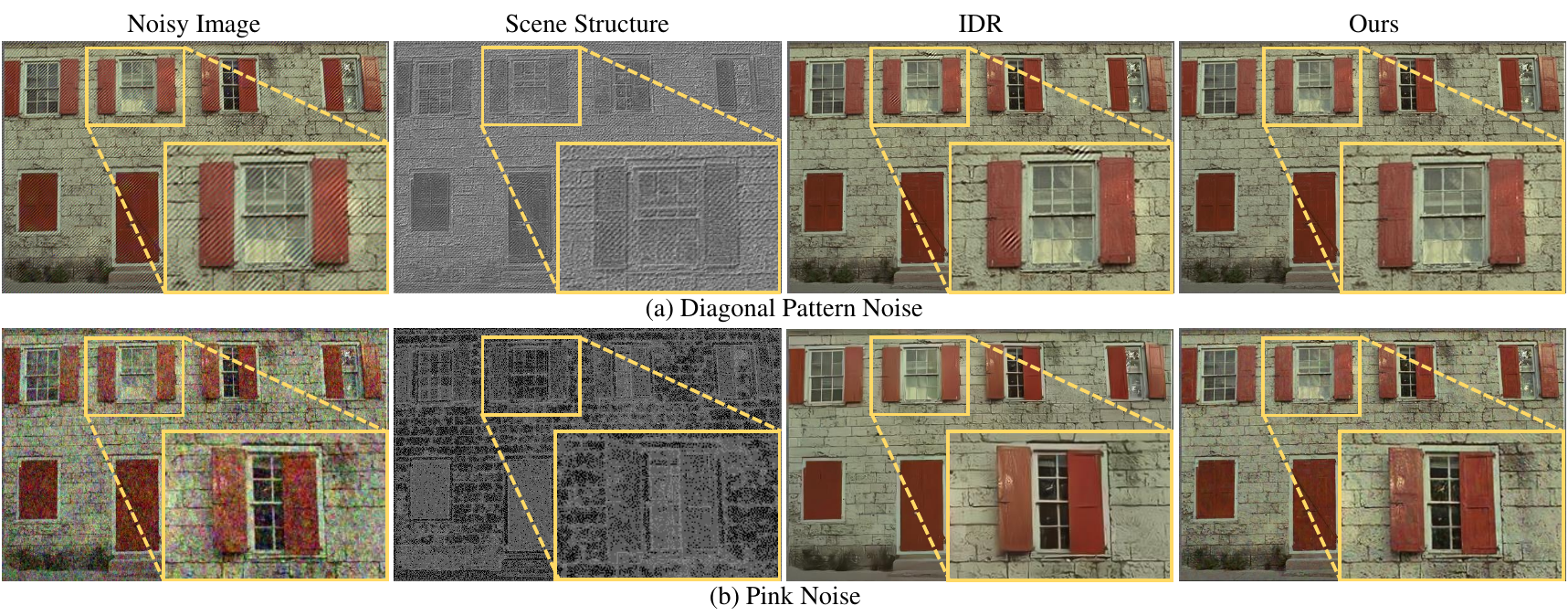}
    \vspace{-6mm}
    \caption{Qualitative comparison for different noise types including diagonal pattern and pink noise with IDR. The structure guidance extracted from the diagonal noisy image tends to extract the diagonal pattern as well as the original structure of the image.}
    \label{fig:noise}
\end{figure*}

\subsubsection{Comparison with Hand-crafted Structure Prior}
Hand-crafted edge detection is widely used for representing scene structures, even in recent models for low-level vision \ie~ inpainting~\citep{guo2021image,dong2022incremental} and super-resolution~\citep{nazeri2019edge}. We employ one of the representative hand-crafted edge map detection methods, Canny edge.

For fair comparison, we generate a set of edge maps with various thresholds for image gradients $\psi$, and embed it into the baseline networks. We note that the edge maps are used as input of our attention layer for the networks to selectively choose informative scene structures during the training phase. Here, we set two types of thresholds $\psi$ to \{0.6, 1.5, 2.4\} and \{1.0, 2.0, 3.0\} that we manually find the best settings to configure image textures and object boundaries.

As shown in \cref{ablation}, the interesting fact is that the performance drop of the joint depth upsampling is not huge when using the hand-crafted edge maps (within 0.07cm in RMSE and 0.037cm in MAE). On the other hand, there is the large performance gap between ours and the Canny edge maps (about 4dB in PSNR and 0.3 in SSIM). 

Two possible reasons why the Canny edge fails to generate task-specific scene representations are: (1) The edge maps are not affected by back-propagation in training phase. 
(2) Based on the experimental results for the image denoising, the Canny edge is sensitive to image noise, which may corrupt the estimated scene structures and eventually not work as a prior. 
On the other hand, as displayed in~\cref{denoising_ablation}, our SSGNet returns the sharpest images, enabling the contents to be read. We thus argue that this experiment demonstrates the efficacy of our learnable structure guidance.

\subsubsection{Other Types of Noises}
Here, we would like to show that SSGNet captures scene structures well in various corruption situations. For this, we conduct an additional denoising experiment with different types of noise w.r.t. spatially correlated noises that utilize convolution kernels to generate pixel-dependent noise, different from a pixel-independent noise like Gaussian noise.
Here, we use both a diagonal line pattern kernel and Gaussian kernel to generate pink noise, following the noise generation scheme of Refined BM3D~\citep{makinen2019exact}. As shown in \cref{tab:noise}, the result demonstrates the overall performance improvement with our SSGNet in terms of PSNR. In case of the diagonal pattern noise, SSGNet tends to extract structures from those patterns, which leads to the relatively low SSIM score. In contrast, SSGNet has a high potential for extracting structures on blurred images with the pink noise. \Cref{fig:noise} also shows the result of each noise type.

\subsubsection{Dataset for initializing SSGNet}
Although we train our SSGNet on NYUv2 for pretraining, we do not have any criteria for selecting pretraining dataset. To show the general performance of our SSGNet regardless to the pretraining dataset, we conduct an experiment on joint depth upsampling with SSGNet pretrained on ImageNet and MIT Adobe 5K. As shown in \cref{tab:dataset}, SSGNet performs consistently well on those dataset.

\subsubsection{Additional Analysis on Parameter Increment}
Although we have shown the performance improvement with our SSGNet embedded to the various baseline networks, it might be questionable whether the performance improvement came from just an increase of parameter size.
Thus, we intentionally increase the number of baseline's parameters and compare the performance. We choose IDR as a baseline image denoising network, following our main experiment.
We tried various changes to the baseline network, such as increasing the channel or kernel size in the convolutional layers. \Cref{tab:comparison} reveals that there is no performance improvement in spite of the increment of network size. 
Worse, the performance drops when the parameter gets larger, which might be caused from overfitting issue. Although larger models have a potential to yield better performance than those with smaller parameters, it needs to make huge efforts to design proper network structures, rather than our plug-and-play scheme.

\begin{table}[t]
\caption{Ablation study for the different types of noise including diagonal pattern and pink noise.}
\centering
\resizebox{0.7\linewidth}{!}{%
\begin{tabular}{c|cccc}
\hline
\multirow{2}{*}{Noise} & \multicolumn{2}{c}{IDR (Baseline)}&\multicolumn{2}{c}{IDR+SSGNet(Ours)}\\
\hhline{|~|----|}
                   &  PSNR & SSIM &PSNR&SSIM\\
\hline
Diagonal Pattern & 41.88 & \textbf{0.993}&\textbf{43.21}&0.987\\
Pink Noise & 28.60 & 0.803&\textbf{30.22}&\textbf{0.818}\\
\hline
\end{tabular}%
}
\label{tab:noise}
\end{table}

\begin{table}[t]
    \caption{Various dataset for initializing SSGNet}
    \centering
    \tiny
    \resizebox{0.7\linewidth}{!}{%
    \begin{tabular}{c|cc|cc}
    \hline
    \multirow{2}{*}{Scale} & \multicolumn{2}{c|}{$\times$4} & \multicolumn{2}{c}{$\times$8} \\
    \hhline{|~|----|}
                           & RMSE & MAE & RMSE & MAE \\
    \hline
    NYUDepthv2 & 0.612 & 0.188 & 0.830 & 0.245\\
    MIT-Adobe 5K & 0.531 & 0.169 & 0.826 & 0.241 \\
    ImageNet & 0.520 & 0.171 & 0.841 & 0.256 \\
    \hline
    \end{tabular}
    }
    \label{tab:dataset}
\end{table}

\begin{table}[t]
\caption{Comparison with initialized SSGNet on the Kodak dataset with noise level $\sigma=50$.}
\centering
\scriptsize
\resizebox{0.7\linewidth}{!}{%
\begin{tabular}{c|ccc}
    \hline
     Models & Parameter Size & PSNR & SSIM \\ 
    \hline
    IDR (Baseline) & 0.99M & 29.27 & 0.803\\
    IDR+SSGNet (Ours) & 1.05M & \textbf{29.34} & \textbf{0.806}\\
    IDR w/ Large Channel & 1.05M & 29.27 & 0.804\\
    IDR w/ Large Kernel & 2.75M & 24.00 & 0.696\\
    \hline
\end{tabular}
}
\label{tab:comparison}
\end{table}

\begin{table*}
\caption{Quantitative results of SSGNet and SSGNet-D on the joint depth upsampling tasks. We use MMSR~\citep{dong2022learning} as baseline network, as an extension of \cref{depthup_sec}.}
\resizebox{\linewidth}{!}{%
\begin{tabular}{c|cccc|cccc|cccc}
\hline
Dataset & \multicolumn{4}{c}{2005} & \multicolumn{4}{|c}{2006} & \multicolumn{4}{|c}{2014} \\
\hline
\multirow{2}{*}{Scale} & \multicolumn{2}{c}{$\times$4} & \multicolumn{2}{c}{$\times$8} & \multicolumn{2}{|c}{$\times$4} & \multicolumn{2}{c}{$\times$8} & \multicolumn{2}{|c}{$\times$4} & \multicolumn{2}{c}{$\times$8} \\
\hhline{|~|------------|}

                     & RMSE & MAE & RMSE & MAE & RMSE & MAE & RMSE & MAE & RMSE & MAE & RMSE & MAE \\
\hline
SSGNet (Ours) & 0.612 & 0.188 & 0.830 & 0.245 & 0.504 & 0.201 & \textbf{0.648} & 0.225 & 1.819 & 0.451 & 2.714 & 0.675 \\
SSGNet-D (Ours$\ast$) & \textbf{0.526} & \textbf{0.177} & \textbf{0.825} & \textbf{0.242} & \textbf{0.459} & \textbf{0.184} & 0.653 & \textbf{0.223} & \textbf{1.710} & \textbf{0.426} & \textbf{2.556} & \textbf{0.642} \\
\hline
\end{tabular}
}
\label{tab:joint_ssgnetd}
\end{table*}

\begin{table*}
\caption{Quantitative results of SSGNet and SSGNet-D on the joint depth upsampling tasks. We use MMSR~\citep{dong2022learning} as baseline network, as an extension of \cref{depthup_sec}.}
\scriptsize
\resizebox{\linewidth}{!}{%
\begin{tabular}{c|c|cc|cc|cc}
\hline
\multirow{2}{*}{Dataset} & \multirow{2}{*}{scale} & \multicolumn{2}{c|}{2005} & \multicolumn{2}{c|}{2006} & \multicolumn{2}{c}{2014}\\
\cline{3-8}
&&RMSE & MAE & RMSE & MAE & RMSE  & MAE\\
\hline
\multirow{2}{*}{SSGNet (Ours)} & $\times$4 & 0.612 & 0.188 & 0.504 & 0.201 & 1.819 & 0.451\\
& $\times$8 & 0.830 & 0.245 & \textbf{0.648} & 0.225 & 2.714 & 0.675\\
\hline
\multirow{2}{*}{SSGNet-D (Ours$\ast$)} & $\times$4 & \textbf{0.526} & \textbf{0.177} & \textbf{0.459} & \textbf{0.184} & \textbf{1.710} & \textbf{0.426} \\ & $\times$8 & \textbf{0.825} & \textbf{0.242} & 0.653 & \textbf{0.223} & \textbf{2.556} & \textbf{0.642}\\
\hline
\end{tabular}
}
\label{tab:joint_ssgnetd2}
\end{table*}

\section{SSGNet-D: Lighter version of SSGNet for Edge Computing Device}
\label{sec:jetson}


The deployment of neural networks on edge computing devices such as smartphones, drones, and medical sensors has gained significant interest.
Since those devices have lack of power and storage, a main concern is how to find an optimal trade-off between accuracy and latency time of models.
Unlike image detection or classification tasks that yield only 2d coordinates for bounding boxes, pixel-wise inference tasks, such as image restoration or depth estimation, require decoders for outputs with the same size of inputs.
Thus, it leads to inevitable performance decrease when the architecture is simplified or compressed for implementing on edge computing devices.

We expect that our structure guidance module can mitigate this performance drop with little latency time on edge devices. Even though our SSGNet already extracts structure guidance information of the scene as well, we focus on finding an optimal solution for edge device and propose a lighter-weight version of SSGNet, SSGNet-D, which is specialized in depth computation tasks.

\begin{table}[t]
\caption{The number of SSGNet-D's parameters and its computational cost compared to SSGNet.}
\centering
\normalsize
\resizebox{0.75\linewidth}{!}{%
\begin{tabular}{c|ccc}
    \hline
    & Parameter Size & MACs & Inference Time\\
    \hline
     SSGNet (Ours) & 56260 & 3.70G & 0.132ms\\
     SSGNet-D (Ours$\ast$) & 28560 & 1.87G & 0.278ms\\
    \hline
\end{tabular}    %
}
\label{tab:ssgnet-d}
\end{table}

\subsection{SSGNet for Depth Computation}
\label{sec:ssgnetd}
To implement on edge computing devices, we further investigate the possibility of lighter-weight module as a structure guidance. 
Image restoration tasks like image denoising or image super-resolution requires capturing fine detailed structures of whole images, including texture and edges. 
However, only the prior information on object boundaries is enough for depth computation with sharp edge boundaries.
Motivated by this fact, we thus propose the reduced size of our SSGNet for depth-related tasks.
\begin{figure}[t]
    \centering
    \includegraphics[clip=true, width=0.7\linewidth]{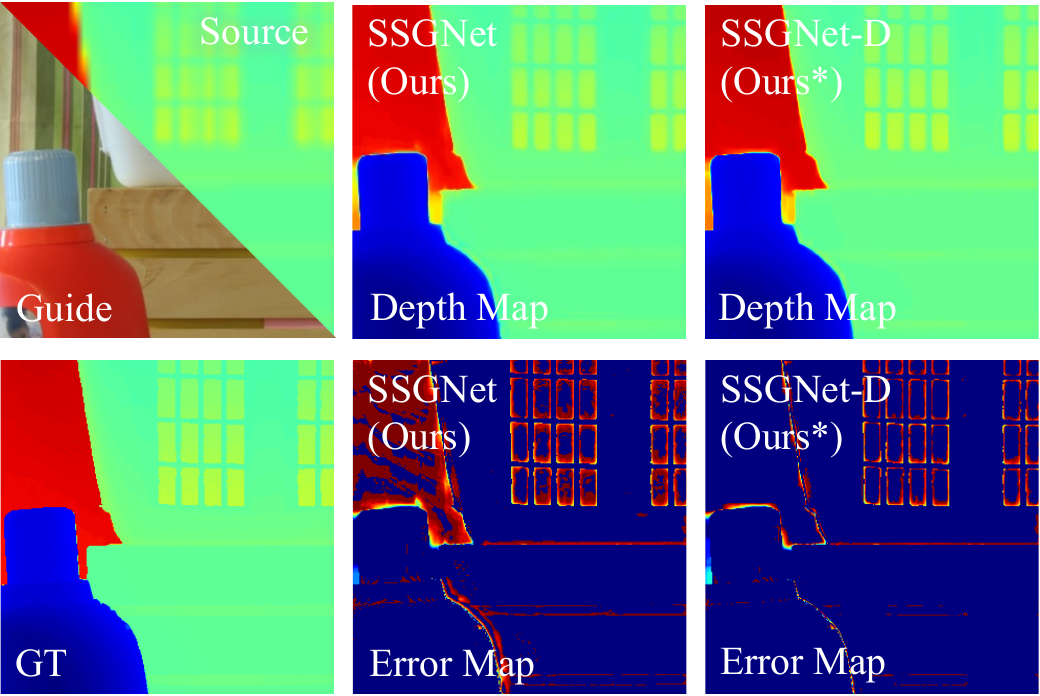}
    \vspace{-5mm}
    \caption{Comparison results of SSGNet and SSGNet-D on the joint depth upsampling with a resolution factor of $8$ on the Middlebury 2005 dataset.}
    \label{fig:joint_depth}
\end{figure}

\begin{figure*}[t]
    \centering
    \includegraphics[clip=true, width=\linewidth]{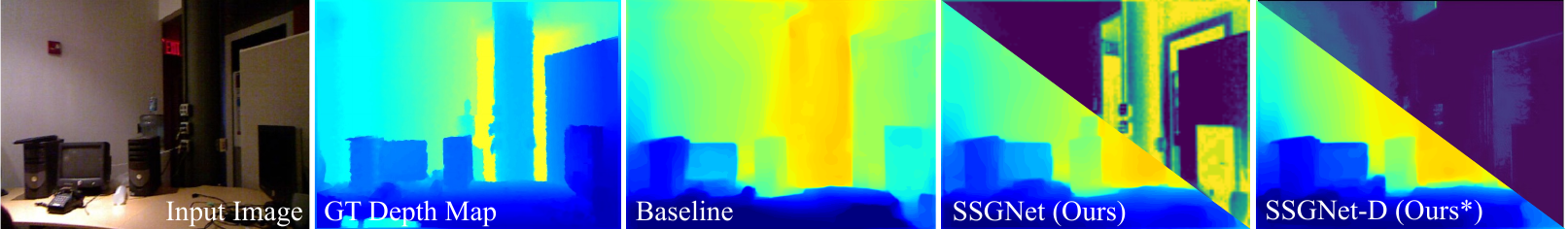}
    \vspace{-5mm}
    \caption{Comparison results on monocular depth estimation on NYUv2 dataset with DenseDepth. We visualize the structure maps extracted from SSGNet and SSGNet-D along with the depth maps. }
    \label{fig:depth_estimation}
\end{figure*}

In order to make the SSGNet much lighter and faster, we bring several structural modifications inside of it.
We adopt a strategy in~\citep{chen2022simple}, which proposes that the nonlinear activation functions like Sigmoid, ReLU, and GELU are not necessary and could be replaced by multiplication or removed.
The work in~\citep{chen2022simple} reviews Gated Linear Units (GLU)~\citep{dauphin2017language} and considers it as a generalization of activation functions.
To simplify the nonlinear activation functions, the work~\citep{chen2022simple} proposes SimpleGate, a simple GLU variant that directly divides feature maps into two parts along with the channel dimension and multiply them.
Given the divided feature maps $A$ and $B$, the SimpleGate is implemented by a simple element-wise multiplication as \cref{eqn:simplegate}:
\begin{equation}
\label{eqn:simplegate}
    \textit{SimpleGate}(A, B) = A\odot B,
\end{equation}
where $\odot$ denotes an element-wise multiplication.
Following the simplification scheme, we replace GELU in all the convolutional layers of SSGNet with SimpleGate.
Since SimpleGate has a half of the channel dimension compared to GELU, we could also significantly reduce the channel size of the convolutions, which are proceeded right after each SimpleGate operation. In addition, we remove the layer normalization from all the convolutional layers to make our SSGNet much faster.

With those modifications, we propose a lighter version of our structure guidance module, SSGNet-D, especially for depth-related tasks. The overall architecture of SSGNet-D is depicted in~\cref{ssgnet} (red colored text), and the number of learnable parameters in SSGNet-D is about 29K.

\begin{figure*}[t]
    \centering
    \includegraphics[clip=true, width=\linewidth]{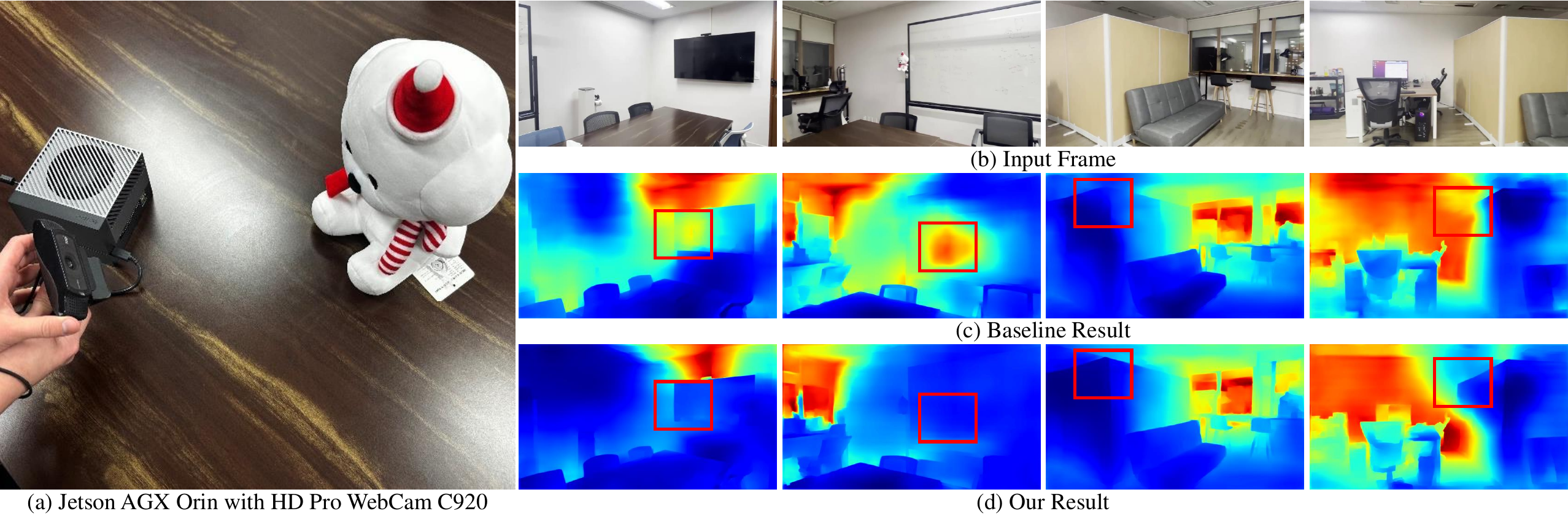}
    \vspace{-5mm}
    \caption{Comparison results on monocular detph estimation implemented on Jetson AGX Orin in the wild environment.}
    \label{fig:jetson}
\end{figure*}

In terms of computational budget, we compute the number of Multiply-ACcumulate (MAC) operator with an input tensor of size (1, 3, 256, 256). 
MAC involves multiplying two values, usually the numbers or vectors, and then adding the result to an accumulator\citep{chen2022simple}.
To compare the performance of SSGNet-D with our original SSGNet, we conduct an additional experiment on joint depth upsampling, as an extension of the experiment conducted previously on \cref{depthup_sec}. Note that we set the same training scheme on SSGNet-D as SSGNet and it took 3 hours to pretrain SSGNet-D, compared to 10 hours for SSGNet.
It is noticeable that the number of MACs coming from SSGNet-D is half of that of SSGNet, as reported in \cref{tab:ssgnet-d} and it works consistently well on joint depth upsampling, even with a fewer parameters ($\sim$ 29K), which is described in \cref{tab:joint_ssgnetd} and \cref{fig:joint_depth}.

We also conduct an additional experiment on monocular depth estimation to show the further applicability of SSGNet on depth computation tasks. Here, we choose DenseDepth~\citep{alhashim2018high} as a baseline, which consists of a simple encoder-decoder architecture with skip connections.
Since they utilize an ImageNet pretrained network as an encoder to extract features from input RGB images, feeding the structure prior along with the image is likely to ruin the internal representation of the encoder.
Thus, we incorporate our structural priors by embedding them just before the final decoder block of convolutional layers as shown in \cref{experiment}(c), accompanied by skip-connections from the encoder, where we use MobileNetv2~\citep{sandler2018mobilenetv2} for further usage in Jetson AGX Orin as a lightweight baseline model. 

We first quantitatively compare the results between the baseline~\citep{alhashim2018high} and ours. Note we train the both from scratch, following the author provided implementation setting. \cref{tab:densedepth} and \cref{fig:depth_estimation} shows that both SSGNet and SSGNet-D improves the performance of baseline model, especially in SSGNet-D. 

Surprisingly, SSGNet-D has the performance gain over the original SSGNet on both experiments in that SSGNet-D is effective to capture the scene structures without any consideration of incidental information of the scene like textures. It is also worth mentioning that our module works well as a structure prior when it is embedded to the decoder of the baseline network.

\subsection{Implementation on edge computing device}
\label{sec:sub_jetson}
To check the performance on edge computing device in the wild, we conduct an experiment using Jetson AGX Orin~\citep{nvidiaorin} for monocular depth estimation in real-world. After pretraining the models to generate onnx file, we convert it to TensorRT, an Nvidia-optimized accelerated library for the learning inference~\citep{nvidiatensorrt}, and run it on Jetson, following the setting of NVIDIA Jetson Project page\footnote{NVIDIA Jetson Deep-learning Project page : \url{https://github.com/dusty-nv/jetson-inference}}. Here, we use HD Pro Webcam C920~\citep{webcam} to capture the in-the-wild scene.

We qualitatively compare the results between the baseline~\citep{alhashim2018high} with and without SSGNet-D. \Cref{fig:jetson} displays some of video sequences in the real-time monocular depth estimation. 
The result demonstrates that our SSGNet contributes to producing improved depth map quality.
Compared to the baseline that takes in average 11.4ms for one depth computation in inference, the baseline with our SSGNet-D take 12.1ms. This fact means that incorporating structural information into the baseline network ensures better real-time performance with little computational delay. Considering that off-the-shelf cameras have at most 60 FPS, it is worth emphasizing that the computational delay does not matter in practice.


\begin{table}[t]
\caption{Additional experiments on NYUv2~\citep{silberman2012indoor} for monocular depth estimation. The results from SSGNet and SSGNet-D are denotes as Ours and Ours$\ast$, respectively. Note that we use MobileNetv2~\citep{sandler2018mobilenetv2} as an encoder for further usage on Jetson AGX Orin.}
\centering
\resizebox{0.7\linewidth}{!}{%
\begin{tabular}{ccccccccccccc}
\hline

Methods & \mcx{2}{|}{$\delta_1\uparrow$} & \mcx{2}{}{$\delta_2\uparrow$} & \mcx{2}{}{$\delta_3\uparrow$} & \mcx{2}{}{rel$\downarrow$} & \mcx{2}{}{rms$\downarrow$} & \mcx{2}{}{$log_{10}\downarrow$}\\
\hline
DenseDepth & \mcx{2}{|}{0.795} & \mcx{2}{}{\underline{0.953}} & \mcx{2}{}{0.987} & \mcx{2}{}{0.152} & \mcx{2}{}{0.534} & \mcx{2}{}{0.063}\\
Ours & \mcx{2}{|}{\underline{0.803}} & \mcx{2}{}{\textbf{0.958}} & \mcx{2}{}{\textbf{0.990}} & \mcx{2}{}{\underline{0.147}} & \mcx{2}{}{\underline{0.521}} & \mcx{2}{}{\underline{0.062}}\\
Ours$\ast$ & \mcx{2}{|}{\textbf{0.808}} & \mcx{2}{}{\textbf{0.958}} & \mcx{2}{}{\underline{0.989}} & \mcx{2}{}{\textbf{0.143}} & \mcx{2}{}{\textbf{0.518}} & \mcx{2}{}{\textbf{0.061}} \\
\hline
\end{tabular}
}
\label{tab:densedepth}
\end{table}

\section{Conclusion}
In this paper, we present a single general network for representing task-specific scene structures.
We cast the problem of the acquisition of informative scene structures as a traditional graph partitioning problem on the image domain, and solve it using a lightweight CNN framework without any supervision, \textit{Scene Structure Guidance Network (SSGNet)}. Our SSGNet computes coefficients of a set of eigenvectors, enabling to efficiently produce diverse feature representations of a scene with a small number of learnable parameters.
With our proposed two loss terms, the eigen loss and the spatial loss, SSGNet is first initialized to parameterize the scene structures. The SSGNet is then embedded into the baseline networks and the parameters are fine-tuned to learn task-specific guidance features as the training proceeds. We show the promising performance gains over off-the-shelf baseline models for various low-level vision tasks such as joint depth upsampling, image denoising, super resolution and low-light image enhancement, even with the good cross-dataset generalization capability. To embed our structural module on edge computing device, which commonly suffer from accuracy-latency trade-off issue, we devise a lighter-weight module, SSGNet-D, depth specialized module. We show that our SSGNet-D improves the performance of the baseline model on Jetson AGX Orin in-the-wild environments, as well as achieving the compatible performance with the original version of SSGNet on depth computation tasks.

\newpage




\bibliographystyle{elsarticle-num-names} 
\bibliography{ref} 

\begin{thebibliography}{90}
\expandafter\ifx\csname natexlab\endcsname\relax\def\natexlab#1{#1}\fi
\providecommand{\url}[1]{\texttt{#1}}
\providecommand{\href}[2]{#2}
\providecommand{\path}[1]{#1}
\providecommand{\DOIprefix}{doi:}
\providecommand{\ArXivprefix}{arXiv:}
\providecommand{\URLprefix}{URL: }
\providecommand{\Pubmedprefix}{pmid:}
\providecommand{\doi}[1]{\href{http://dx.doi.org/#1}{\path{#1}}}
\providecommand{\Pubmed}[1]{\href{pmid:#1}{\path{#1}}}
\providecommand{\bibinfo}[2]{#2}
\ifx\xfnm\relax \def\xfnm[#1]{\unskip,\space#1}\fi
\bibitem[{Tomasi and Manduchi(1998)}]{tomasi1998bilateral}
\bibinfo{author}{C.~Tomasi}, \bibinfo{author}{R.~Manduchi},
\newblock \bibinfo{title}{Bilateral filtering for gray and color images},
\newblock in: \bibinfo{booktitle}{Proceedings of International Conference on Computer Vision (ICCV)}, \bibinfo{year}{1998}.
\bibitem[{Krishnan and Fergus(2009)}]{krishnan2009fast}
\bibinfo{author}{D.~Krishnan}, \bibinfo{author}{R.~Fergus},
\newblock \bibinfo{title}{Fast image deconvolution using hyper-laplacian priors},
\newblock in: \bibinfo{booktitle}{Proceedings of the Neural Information Processing Systems (NeurIPS)}, \bibinfo{year}{2009}.
\bibitem[{Levin et~al.(2007)Levin, Fergus, Durand, and Freeman}]{levin2007image}
\bibinfo{author}{A.~Levin}, \bibinfo{author}{R.~Fergus}, \bibinfo{author}{F.~Durand}, \bibinfo{author}{W.~T. Freeman},
\newblock \bibinfo{title}{Image and depth from a conventional camera with a coded aperture},
\newblock \bibinfo{journal}{ACM transactions on graphics (TOG)} \bibinfo{volume}{26} (\bibinfo{year}{2007}) \bibinfo{pages}{70--es}.
\bibitem[{Tai et~al.(2010)Tai, Liu, Brown, and Lin}]{tai2010super}
\bibinfo{author}{Y.-W. Tai}, \bibinfo{author}{S.~Liu}, \bibinfo{author}{M.~S. Brown}, \bibinfo{author}{S.~Lin},
\newblock \bibinfo{title}{Super resolution using edge prior and single image detail synthesis},
\newblock in: \bibinfo{booktitle}{Proceedings of IEEE Conference on Computer Vision and Pattern Recognition (CVPR)}, \bibinfo{year}{2010}.
\bibitem[{Nazeri et~al.(2019)Nazeri, Ng, Joseph, Qureshi, and Ebrahimi}]{nazeri2019edgeconnect}
\bibinfo{author}{K.~Nazeri}, \bibinfo{author}{E.~Ng}, \bibinfo{author}{T.~Joseph}, \bibinfo{author}{F.~Qureshi}, \bibinfo{author}{M.~Ebrahimi},
\newblock \bibinfo{title}{Edgeconnect: Structure guided image inpainting using edge prediction},
\newblock in: \bibinfo{booktitle}{Proceedings of International Conference on Computer Vision Workshop (ICCVW)}, \bibinfo{year}{2019}.
\bibitem[{Yang et~al.(2020)Yang, Qi, and Shi}]{yang2020learning}
\bibinfo{author}{J.~Yang}, \bibinfo{author}{Z.~Qi}, \bibinfo{author}{Y.~Shi},
\newblock \bibinfo{title}{Learning to incorporate structure knowledge for image inpainting},
\newblock in: \bibinfo{booktitle}{Proceedings of the AAAI Conference on Artificial Intelligence (AAAI)}, \bibinfo{year}{2020}.
\bibitem[{Guo et~al.(2021)Guo, Yang, and Huang}]{guo2021image}
\bibinfo{author}{X.~Guo}, \bibinfo{author}{H.~Yang}, \bibinfo{author}{D.~Huang},
\newblock \bibinfo{title}{Image inpainting via conditional texture and structure dual generation},
\newblock in: \bibinfo{booktitle}{Proceedings of International Conference on Computer Vision (ICCV)}, \bibinfo{year}{2021}.
\bibitem[{He et~al.(2012)He, Sun, and Tang}]{he2012guided}
\bibinfo{author}{K.~He}, \bibinfo{author}{J.~Sun}, \bibinfo{author}{X.~Tang},
\newblock \bibinfo{title}{Guided image filtering},
\newblock \bibinfo{journal}{IEEE Transactions on Pattern Analysis and Machine Intelligence (TPAMI)} \bibinfo{volume}{35} (\bibinfo{year}{2012}) \bibinfo{pages}{1397--1409}.
\bibitem[{Guo et~al.(2018)Guo, Li, Ma, and Ling}]{guo2018mutually}
\bibinfo{author}{X.~Guo}, \bibinfo{author}{Y.~Li}, \bibinfo{author}{J.~Ma}, \bibinfo{author}{H.~Ling},
\newblock \bibinfo{title}{Mutually guided image filtering},
\newblock \bibinfo{journal}{IEEE Transactions on Pattern Analysis and Machine Intelligence (TPAMI)} \bibinfo{volume}{42} (\bibinfo{year}{2018}) \bibinfo{pages}{694--707}.
\bibitem[{Li et~al.(2016)Li, Huang, Ahuja, and Yang}]{li2016deep}
\bibinfo{author}{Y.~Li}, \bibinfo{author}{J.-B. Huang}, \bibinfo{author}{N.~Ahuja}, \bibinfo{author}{M.-H. Yang},
\newblock \bibinfo{title}{Deep joint image filtering},
\newblock in: \bibinfo{booktitle}{Proceedings of European Conference on Computer Vision (ECCV)}, \bibinfo{year}{2016}.
\bibitem[{Eldesokey et~al.(2020)Eldesokey, Felsberg, Holmquist, and Persson}]{eldesokey2020uncertainty}
\bibinfo{author}{A.~Eldesokey}, \bibinfo{author}{M.~Felsberg}, \bibinfo{author}{K.~Holmquist}, \bibinfo{author}{M.~Persson},
\newblock \bibinfo{title}{Uncertainty-aware cnns for depth completion: Uncertainty from beginning to end},
\newblock in: \bibinfo{booktitle}{Proceedings of IEEE Conference on Computer Vision and Pattern Recognition (CVPR)}, \bibinfo{year}{2020}.
\bibitem[{de~Lutio et~al.(2022)de~Lutio, Becker, D'Aronco, Russo, Wegner, and Schindler}]{de2022learning}
\bibinfo{author}{R.~de~Lutio}, \bibinfo{author}{A.~Becker}, \bibinfo{author}{S.~D'Aronco}, \bibinfo{author}{S.~Russo}, \bibinfo{author}{J.~D. Wegner}, \bibinfo{author}{K.~Schindler},
\newblock \bibinfo{title}{Learning graph regularisation for guided super-resolution},
\newblock in: \bibinfo{booktitle}{Proceedings of IEEE Conference on Computer Vision and Pattern Recognition (CVPR)}, \bibinfo{year}{2022}.
\bibitem[{Irwin et~al.(1968)}]{irwin1968isotropic}
\bibinfo{author}{F.~Irwin}, et~al.,
\newblock \bibinfo{title}{An isotropic 3x3 image gradient operator},
\newblock \bibinfo{journal}{Presentation at Stanford AI Project} \bibinfo{volume}{2014} (\bibinfo{year}{1968}).
\bibitem[{Canny(1986)}]{canny1986computational}
\bibinfo{author}{J.~Canny},
\newblock \bibinfo{title}{A computational approach to edge detection},
\newblock \bibinfo{journal}{IEEE Transactions on Pattern Analysis and Machine Intelligence (TPAMI)} \bibinfo{volume}{8} (\bibinfo{year}{1986}) \bibinfo{pages}{679--698}.
\bibitem[{Fang et~al.(2020)Fang, Li, and Zeng}]{fang2020soft}
\bibinfo{author}{F.~Fang}, \bibinfo{author}{J.~Li}, \bibinfo{author}{T.~Zeng},
\newblock \bibinfo{title}{Soft-edge assisted network for single image super-resolution},
\newblock \bibinfo{journal}{IEEE Transactions on Image Processing (TIP)} \bibinfo{volume}{29} (\bibinfo{year}{2020}) \bibinfo{pages}{4656--4668}.
\bibitem[{Shi and Malik(2000)}]{shi2000normalized}
\bibinfo{author}{J.~Shi}, \bibinfo{author}{J.~Malik},
\newblock \bibinfo{title}{Normalized cuts and image segmentation},
\newblock \bibinfo{journal}{IEEE Transactions on Pattern Analysis and Machine Intelligence (TPAMI)} \bibinfo{volume}{22} (\bibinfo{year}{2000}) \bibinfo{pages}{888--905}.
\bibitem[{Cohen and Shashua(2017)}]{cohen2016inductive}
\bibinfo{author}{N.~Cohen}, \bibinfo{author}{A.~Shashua},
\newblock \bibinfo{title}{Inductive bias of deep convolutional networks through pooling geometry},
\newblock in: \bibinfo{booktitle}{International Conference on Learning Representations (ICLR)}, \bibinfo{year}{2017}.
\bibitem[{Zhang and Sabuncu(2018)}]{zhang2018generalized}
\bibinfo{author}{Z.~Zhang}, \bibinfo{author}{M.~Sabuncu},
\newblock \bibinfo{title}{Generalized cross entropy loss for training deep neural networks with noisy labels},
\newblock in: \bibinfo{booktitle}{Proceedings of the Neural Information Processing Systems (NeurIPS)}, \bibinfo{year}{2018}.
\bibitem[{Zhong et~al.(2021)Zhong, Yuan, Wu, Yuan, Peng, and Wang}]{zhong2021pixel}
\bibinfo{author}{Y.~Zhong}, \bibinfo{author}{B.~Yuan}, \bibinfo{author}{H.~Wu}, \bibinfo{author}{Z.~Yuan}, \bibinfo{author}{J.~Peng}, \bibinfo{author}{Y.-X. Wang},
\newblock \bibinfo{title}{Pixel contrastive-consistent semi-supervised semantic segmentation},
\newblock in: \bibinfo{booktitle}{Proceedings of International Conference on Computer Vision (ICCV)}, \bibinfo{year}{2021}.
\bibitem[{Goodfellow et~al.(2014)Goodfellow, Pouget-Abadie, Mirza, Xu, Warde-Farley, Ozair, Courville, and Bengio}]{goodfellow2014generative}
\bibinfo{author}{I.~Goodfellow}, \bibinfo{author}{J.~Pouget-Abadie}, \bibinfo{author}{M.~Mirza}, \bibinfo{author}{B.~Xu}, \bibinfo{author}{D.~Warde-Farley}, \bibinfo{author}{S.~Ozair}, \bibinfo{author}{A.~Courville}, \bibinfo{author}{Y.~Bengio},
\newblock \bibinfo{title}{Generative adversarial nets},
\newblock in: \bibinfo{booktitle}{Proceedings of the Neural Information Processing Systems (NeurIPS)}, \bibinfo{year}{2014}.
\bibitem[{Zhu et~al.(2017)Zhu, Park, Isola, and Efros}]{zhu2017unpaired}
\bibinfo{author}{J.-Y. Zhu}, \bibinfo{author}{T.~Park}, \bibinfo{author}{P.~Isola}, \bibinfo{author}{A.~A. Efros},
\newblock \bibinfo{title}{Unpaired image-to-image translation using cycle-consistent adversarial networks},
\newblock in: \bibinfo{booktitle}{Proceedings of International Conference on Computer Vision (ICCV)}, \bibinfo{year}{2017}.
\bibitem[{Karras et~al.(2019)Karras, Laine, and Aila}]{karras2019style}
\bibinfo{author}{T.~Karras}, \bibinfo{author}{S.~Laine}, \bibinfo{author}{T.~Aila},
\newblock \bibinfo{title}{A style-based generator architecture for generative adversarial networks},
\newblock in: \bibinfo{booktitle}{Proceedings of IEEE Conference on Computer Vision and Pattern Recognition (CVPR)}, \bibinfo{year}{2019}.
\bibitem[{Liu et~al.(2021)Liu, Wan, Huang, Song, Han, and Liao}]{liu2021pd}
\bibinfo{author}{H.~Liu}, \bibinfo{author}{Z.~Wan}, \bibinfo{author}{W.~Huang}, \bibinfo{author}{Y.~Song}, \bibinfo{author}{X.~Han}, \bibinfo{author}{J.~Liao},
\newblock \bibinfo{title}{Pd-gan: Probabilistic diverse gan for image inpainting},
\newblock in: \bibinfo{booktitle}{Proceedings of IEEE Conference on Computer Vision and Pattern Recognition (CVPR)}, \bibinfo{year}{2021}.
\bibitem[{Wang et~al.(2021)Wang, Xie, Dong, and Shan}]{wang2021real}
\bibinfo{author}{X.~Wang}, \bibinfo{author}{L.~Xie}, \bibinfo{author}{C.~Dong}, \bibinfo{author}{Y.~Shan},
\newblock \bibinfo{title}{Real-esrgan: Training real-world blind super-resolution with pure synthetic data},
\newblock in: \bibinfo{booktitle}{Proceedings of International Conference on Computer Vision (ICCV)}, \bibinfo{year}{2021}.
\bibitem[{Johnson et~al.(2016)Johnson, Alahi, and Fei-Fei}]{johnson2016perceptual}
\bibinfo{author}{J.~Johnson}, \bibinfo{author}{A.~Alahi}, \bibinfo{author}{L.~Fei-Fei},
\newblock \bibinfo{title}{Perceptual losses for real-time style transfer and super-resolution},
\newblock in: \bibinfo{booktitle}{Proceedings of European Conference on Computer Vision (ECCV)}, \bibinfo{year}{2016}.
\bibitem[{Fuoli et~al.(2021)Fuoli, Van~Gool, and Timofte}]{fuoli2021fourier}
\bibinfo{author}{D.~Fuoli}, \bibinfo{author}{L.~Van~Gool}, \bibinfo{author}{R.~Timofte},
\newblock \bibinfo{title}{Fourier space losses for efficient perceptual image super-resolution},
\newblock in: \bibinfo{booktitle}{Proceedings of International Conference on Computer Vision (ICCV)}, \bibinfo{year}{2021}.
\bibitem[{Suvorov et~al.(2022)Suvorov, Logacheva, Mashikhin, Remizova, Ashukha, Silvestrov, Kong, Goka, Park, and Lempitsky}]{suvorov2022resolution}
\bibinfo{author}{R.~Suvorov}, \bibinfo{author}{E.~Logacheva}, \bibinfo{author}{A.~Mashikhin}, \bibinfo{author}{A.~Remizova}, \bibinfo{author}{A.~Ashukha}, \bibinfo{author}{A.~Silvestrov}, \bibinfo{author}{N.~Kong}, \bibinfo{author}{H.~Goka}, \bibinfo{author}{K.~Park}, \bibinfo{author}{V.~Lempitsky},
\newblock \bibinfo{title}{Resolution-robust large mask inpainting with fourier convolutions},
\newblock in: \bibinfo{booktitle}{Proceedings of the IEEE/CVF Winter Conference on Applications of Computer Vision (WACV)}, \bibinfo{year}{2022}.
\bibitem[{Dosovitskiy et~al.(2021)Dosovitskiy, Beyer, Kolesnikov, Weissenborn, Zhai, Unterthiner, Dehghani, Minderer, Heigold, Gelly et~al.}]{dosovitskiy2020image}
\bibinfo{author}{A.~Dosovitskiy}, \bibinfo{author}{L.~Beyer}, \bibinfo{author}{A.~Kolesnikov}, \bibinfo{author}{D.~Weissenborn}, \bibinfo{author}{X.~Zhai}, \bibinfo{author}{T.~Unterthiner}, \bibinfo{author}{M.~Dehghani}, \bibinfo{author}{M.~Minderer}, \bibinfo{author}{G.~Heigold}, \bibinfo{author}{S.~Gelly}, et~al.,
\newblock \bibinfo{title}{An image is worth 16x16 words: Transformers for image recognition at scale},
\newblock in: \bibinfo{booktitle}{International Conference on Learning Representations (ICLR)}, \bibinfo{year}{2021}.
\bibitem[{Liu et~al.(2021)Liu, Lin, Cao, Hu, Wei, Zhang, Lin, and Guo}]{liu2021swin}
\bibinfo{author}{Z.~Liu}, \bibinfo{author}{Y.~Lin}, \bibinfo{author}{Y.~Cao}, \bibinfo{author}{H.~Hu}, \bibinfo{author}{Y.~Wei}, \bibinfo{author}{Z.~Zhang}, \bibinfo{author}{S.~Lin}, \bibinfo{author}{B.~Guo},
\newblock \bibinfo{title}{Swin transformer: Hierarchical vision transformer using shifted windows},
\newblock in: \bibinfo{booktitle}{Proceedings of International Conference on Computer Vision (ICCV)}, \bibinfo{year}{2021}.
\bibitem[{Caron et~al.(2021)Caron, Touvron, Misra, J{\'e}gou, Mairal, Bojanowski, and Joulin}]{caron2021emerging}
\bibinfo{author}{M.~Caron}, \bibinfo{author}{H.~Touvron}, \bibinfo{author}{I.~Misra}, \bibinfo{author}{H.~J{\'e}gou}, \bibinfo{author}{J.~Mairal}, \bibinfo{author}{P.~Bojanowski}, \bibinfo{author}{A.~Joulin},
\newblock \bibinfo{title}{Emerging properties in self-supervised vision transformers},
\newblock in: \bibinfo{booktitle}{Proceedings of International Conference on Computer Vision (ICCV)}, \bibinfo{year}{2021}.
\bibitem[{Chen et~al.(2021)Chen, Wang, Guo, Xu, Deng, Liu, Ma, Xu, Xu, and Gao}]{chen2021pre}
\bibinfo{author}{H.~Chen}, \bibinfo{author}{Y.~Wang}, \bibinfo{author}{T.~Guo}, \bibinfo{author}{C.~Xu}, \bibinfo{author}{Y.~Deng}, \bibinfo{author}{Z.~Liu}, \bibinfo{author}{S.~Ma}, \bibinfo{author}{C.~Xu}, \bibinfo{author}{C.~Xu}, \bibinfo{author}{W.~Gao},
\newblock \bibinfo{title}{Pre-trained image processing transformer},
\newblock in: \bibinfo{booktitle}{Proceedings of IEEE Conference on Computer Vision and Pattern Recognition (CVPR)}, \bibinfo{year}{2021}.
\bibitem[{Tibshirani(1996)}]{tibshirani1996regression}
\bibinfo{author}{R.~Tibshirani},
\newblock \bibinfo{title}{Regression shrinkage and selection via the lasso},
\newblock \bibinfo{journal}{Journal of the Royal Statistical Society: Series B (Methodological)} \bibinfo{volume}{58} (\bibinfo{year}{1996}) \bibinfo{pages}{267--288}.
\bibitem[{Xu et~al.(2010)Xu, Zhang, Wang, Chang, and Liang}]{xu2010l1}
\bibinfo{author}{Z.~Xu}, \bibinfo{author}{H.~Zhang}, \bibinfo{author}{Y.~Wang}, \bibinfo{author}{X.~Chang}, \bibinfo{author}{Y.~Liang},
\newblock \bibinfo{title}{L1/2 regularization},
\newblock \bibinfo{journal}{Science China Information Sciences} \bibinfo{volume}{53} (\bibinfo{year}{2010}) \bibinfo{pages}{1159--1169}.
\bibitem[{Loshchilov and Hutter(2019)}]{loshchilov2017decoupled}
\bibinfo{author}{I.~Loshchilov}, \bibinfo{author}{F.~Hutter},
\newblock \bibinfo{title}{Decoupled weight decay regularization},
\newblock in: \bibinfo{booktitle}{International Conference on Learning Representations (ICLR)}, \bibinfo{year}{2019}.
\bibitem[{Zhang et~al.(2014)Zhang, Shen, Xu, and Jia}]{zhang2014rolling}
\bibinfo{author}{Q.~Zhang}, \bibinfo{author}{X.~Shen}, \bibinfo{author}{L.~Xu}, \bibinfo{author}{J.~Jia},
\newblock \bibinfo{title}{Rolling guidance filter},
\newblock in: \bibinfo{booktitle}{Proceedings of European Conference on Computer Vision (ECCV)}, \bibinfo{year}{2014}.
\bibitem[{Pickup et~al.(2003)Pickup, Roberts, and Zisserman}]{pickup2003sampled}
\bibinfo{author}{L.~Pickup}, \bibinfo{author}{S.~J. Roberts}, \bibinfo{author}{A.~Zisserman},
\newblock \bibinfo{title}{A sampled texture prior for image super-resolution},
\newblock in: \bibinfo{booktitle}{Proceedings of the Neural Information Processing Systems (NeurIPS)}, \bibinfo{year}{2003}.
\bibitem[{Sun et~al.(2008)Sun, Xu, and Shum}]{sun2008image}
\bibinfo{author}{J.~Sun}, \bibinfo{author}{Z.~Xu}, \bibinfo{author}{H.-Y. Shum},
\newblock \bibinfo{title}{Image super-resolution using gradient profile prior},
\newblock in: \bibinfo{booktitle}{Proceedings of IEEE Conference on Computer Vision and Pattern Recognition (CVPR)}, \bibinfo{year}{2008}.
\bibitem[{Xie et~al.(2015)Xie, Feris, and Sun}]{xie2015edge}
\bibinfo{author}{J.~Xie}, \bibinfo{author}{R.~S. Feris}, \bibinfo{author}{M.-T. Sun},
\newblock \bibinfo{title}{Edge-guided single depth image super resolution},
\newblock \bibinfo{journal}{IEEE Transactions on Image Processing (TIP)} \bibinfo{volume}{25} (\bibinfo{year}{2015}) \bibinfo{pages}{428--438}.
\bibitem[{Liu et~al.(2020)Liu, Anwar, Zheng, and Tian}]{liu2020gradnet}
\bibinfo{author}{Y.~Liu}, \bibinfo{author}{S.~Anwar}, \bibinfo{author}{L.~Zheng}, \bibinfo{author}{Q.~Tian},
\newblock \bibinfo{title}{Gradnet image denoising},
\newblock in: \bibinfo{booktitle}{Proceedings of IEEE Conference on Computer Vision and Pattern Recognition Workshop (CVPRW)}, \bibinfo{year}{2020}.
\bibitem[{Gu et~al.(2017)Gu, Zuo, Guo, Chen, Chen, and Zhang}]{gu2017learning}
\bibinfo{author}{S.~Gu}, \bibinfo{author}{W.~Zuo}, \bibinfo{author}{S.~Guo}, \bibinfo{author}{Y.~Chen}, \bibinfo{author}{C.~Chen}, \bibinfo{author}{L.~Zhang},
\newblock \bibinfo{title}{Learning dynamic guidance for depth image enhancement},
\newblock in: \bibinfo{booktitle}{Proceedings of IEEE Conference on Computer Vision and Pattern Recognition (CVPR)}, \bibinfo{year}{2017}.
\bibitem[{Jin et~al.(2020)Jin, Xu, Zheng, Zhang, Tang, Xu, Yu, and Gao}]{jin2020geometric}
\bibinfo{author}{L.~Jin}, \bibinfo{author}{Y.~Xu}, \bibinfo{author}{J.~Zheng}, \bibinfo{author}{J.~Zhang}, \bibinfo{author}{R.~Tang}, \bibinfo{author}{S.~Xu}, \bibinfo{author}{J.~Yu}, \bibinfo{author}{S.~Gao},
\newblock \bibinfo{title}{Geometric structure based and regularized depth estimation from 360 indoor imagery},
\newblock in: \bibinfo{booktitle}{Proceedings of IEEE Conference on Computer Vision and Pattern Recognition (CVPR)}, \bibinfo{year}{2020}.
\bibitem[{Cao and Fu(2021)}]{cao2021learning}
\bibinfo{author}{C.~Cao}, \bibinfo{author}{Y.~Fu},
\newblock \bibinfo{title}{Learning a sketch tensor space for image inpainting of man-made scenes},
\newblock in: \bibinfo{booktitle}{Proceedings of International Conference on Computer Vision (ICCV)}, \bibinfo{year}{2021}.
\bibitem[{Levin et~al.(2008)Levin, Rav-Acha, and Lischinski}]{levin2008spectral}
\bibinfo{author}{A.~Levin}, \bibinfo{author}{A.~Rav-Acha}, \bibinfo{author}{D.~Lischinski},
\newblock \bibinfo{title}{Spectral matting},
\newblock \bibinfo{journal}{IEEE Transactions on Pattern Analysis and Machine Intelligence (TPAMI)} \bibinfo{volume}{30} (\bibinfo{year}{2008}) \bibinfo{pages}{1699--1712}.
\bibitem[{Levin et~al.(2007)Levin, Lischinski, and Weiss}]{levin2007closed}
\bibinfo{author}{A.~Levin}, \bibinfo{author}{D.~Lischinski}, \bibinfo{author}{Y.~Weiss},
\newblock \bibinfo{title}{A closed-form solution to natural image matting},
\newblock \bibinfo{journal}{IEEE Transactions on Pattern Analysis and Machine Intelligence (TPAMI)} \bibinfo{volume}{30} (\bibinfo{year}{2007}) \bibinfo{pages}{228--242}.
\bibitem[{Tang and Tan(2019)}]{tang2018ba}
\bibinfo{author}{C.~Tang}, \bibinfo{author}{P.~Tan},
\newblock \bibinfo{title}{Ba-net: Dense bundle adjustment network},
\newblock in: \bibinfo{booktitle}{International Conference on Learning Representations (ICLR)}, \bibinfo{year}{2019}.
\bibitem[{Bloesch et~al.(2018)Bloesch, Czarnowski, Clark, Leutenegger, and Davison}]{bloesch2018codeslam}
\bibinfo{author}{M.~Bloesch}, \bibinfo{author}{J.~Czarnowski}, \bibinfo{author}{R.~Clark}, \bibinfo{author}{S.~Leutenegger}, \bibinfo{author}{A.~J. Davison},
\newblock \bibinfo{title}{Codeslam—learning a compact, optimisable representation for dense visual slam},
\newblock in: \bibinfo{booktitle}{Proceedings of IEEE Conference on Computer Vision and Pattern Recognition (CVPR)}, \bibinfo{year}{2018}.
\bibitem[{Fiedler(1973)}]{fiedler1973algebraic}
\bibinfo{author}{M.~Fiedler},
\newblock \bibinfo{title}{Algebraic connectivity of graphs},
\newblock \bibinfo{journal}{Czechoslovak mathematical journal} \bibinfo{volume}{23} (\bibinfo{year}{1973}) \bibinfo{pages}{298--305}.
\bibitem[{Boykov et~al.(2001)Boykov, Veksler, and Zabih}]{boykov2001fast}
\bibinfo{author}{Y.~Boykov}, \bibinfo{author}{O.~Veksler}, \bibinfo{author}{R.~Zabih},
\newblock \bibinfo{title}{Fast approximate energy minimization via graph cuts},
\newblock \bibinfo{journal}{IEEE Transactions on Pattern Analysis and Machine Intelligence (TPAMI)} \bibinfo{volume}{23} (\bibinfo{year}{2001}) \bibinfo{pages}{1222--1239}.
\bibitem[{Ba et~al.(2016)Ba, Kiros, and Hinton}]{ba2016layer}
\bibinfo{author}{J.~L. Ba}, \bibinfo{author}{J.~R. Kiros}, \bibinfo{author}{G.~E. Hinton}, \bibinfo{title}{Layer normalization}, \bibinfo{year}{2016}.
\bibitem[{Hendrycks and Gimpel(2016)}]{hendrycks2016gaussian}
\bibinfo{author}{D.~Hendrycks}, \bibinfo{author}{K.~Gimpel}, \bibinfo{title}{Gaussian error linear units (gelus)}, \bibinfo{year}{2016}.
\bibitem[{Chen et~al.(2013)Chen, Li, and Tang}]{chen2013knn}
\bibinfo{author}{Q.~Chen}, \bibinfo{author}{D.~Li}, \bibinfo{author}{C.-K. Tang},
\newblock \bibinfo{title}{Knn matting},
\newblock \bibinfo{journal}{IEEE Transactions on Pattern Analysis and Machine Intelligence (TPAMI)} \bibinfo{volume}{35} (\bibinfo{year}{2013}) \bibinfo{pages}{2175--2188}.
\bibitem[{Cover and Hart(1967)}]{cover1967nearest}
\bibinfo{author}{T.~Cover}, \bibinfo{author}{P.~Hart},
\newblock \bibinfo{title}{Nearest neighbor pattern classification},
\newblock \bibinfo{journal}{IEEE Transactions on Information Theory} \bibinfo{volume}{13} (\bibinfo{year}{1967}) \bibinfo{pages}{21--27}.
\bibitem[{Dong et~al.(2022)Dong, Yokoya, Wang, and Uezato}]{dong2022learning}
\bibinfo{author}{X.~Dong}, \bibinfo{author}{N.~Yokoya}, \bibinfo{author}{L.~Wang}, \bibinfo{author}{T.~Uezato},
\newblock \bibinfo{title}{Learning mutual modulation for self-supervised cross-modal super-resolution},
\newblock in: \bibinfo{booktitle}{Proceedings of European Conference on Computer Vision (ECCV)}, \bibinfo{year}{2022}.
\bibitem[{Zhang et~al.(2022)Zhang, Li, Law, Wang, Qin, and Li}]{zhang2022idr}
\bibinfo{author}{Y.~Zhang}, \bibinfo{author}{D.~Li}, \bibinfo{author}{K.~L. Law}, \bibinfo{author}{X.~Wang}, \bibinfo{author}{H.~Qin}, \bibinfo{author}{H.~Li},
\newblock \bibinfo{title}{Idr: Self-supervised image denoising via iterative data refinement},
\newblock in: \bibinfo{booktitle}{Proceedings of IEEE Conference on Computer Vision and Pattern Recognition (CVPR)}, \bibinfo{year}{2022}.
\bibitem[{Li et~al.(2014)Li, Lu, Zeng, Gan, and Zha}]{li2014similarity}
\bibinfo{author}{J.~Li}, \bibinfo{author}{Z.~Lu}, \bibinfo{author}{G.~Zeng}, \bibinfo{author}{R.~Gan}, \bibinfo{author}{H.~Zha},
\newblock \bibinfo{title}{Similarity-aware patchwork assembly for depth image super-resolution},
\newblock in: \bibinfo{booktitle}{Proceedings of IEEE Conference on Computer Vision and Pattern Recognition (CVPR)}, \bibinfo{year}{2014}.
\bibitem[{LeCun et~al.(1989)LeCun, Boser, Denker, Henderson, Howard, Hubbard, and Jackel}]{lecun1989backpropagation}
\bibinfo{author}{Y.~LeCun}, \bibinfo{author}{B.~Boser}, \bibinfo{author}{J.~S. Denker}, \bibinfo{author}{D.~Henderson}, \bibinfo{author}{R.~E. Howard}, \bibinfo{author}{W.~Hubbard}, \bibinfo{author}{L.~D. Jackel},
\newblock \bibinfo{title}{Backpropagation applied to handwritten zip code recognition},
\newblock \bibinfo{journal}{Neural computation} \bibinfo{volume}{1} (\bibinfo{year}{1989}) \bibinfo{pages}{541--551}.
\bibitem[{Paszke et~al.(2019)Paszke, Gross, Massa, Lerer, Bradbury, Chanan, Killeen, Lin, Gimelshein, Antiga et~al.}]{paszke2019pytorch}
\bibinfo{author}{A.~Paszke}, \bibinfo{author}{S.~Gross}, \bibinfo{author}{F.~Massa}, \bibinfo{author}{A.~Lerer}, \bibinfo{author}{J.~Bradbury}, \bibinfo{author}{G.~Chanan}, \bibinfo{author}{T.~Killeen}, \bibinfo{author}{Z.~Lin}, \bibinfo{author}{N.~Gimelshein}, \bibinfo{author}{L.~Antiga}, et~al.,
\newblock \bibinfo{title}{Pytorch: An imperative style, high-performance deep learning library},
\newblock in: \bibinfo{booktitle}{Proceedings of the Neural Information Processing Systems (NeurIPS)}, \bibinfo{year}{2019}.
\bibitem[{Kingma and Ba(2014)}]{kingma2014adam}
\bibinfo{author}{D.~P. Kingma}, \bibinfo{author}{J.~Ba},
\newblock \bibinfo{title}{Adam: A method for stochastic optimization},
\newblock in: \bibinfo{booktitle}{International Conference on Learning Representations (ICLR)}, \bibinfo{year}{2014}.
\bibitem[{Silberman and Fergus(2011)}]{silberman2011indoor}
\bibinfo{author}{N.~Silberman}, \bibinfo{author}{R.~Fergus},
\newblock \bibinfo{title}{Indoor scene segmentation using a structured light sensor},
\newblock in: \bibinfo{booktitle}{Proceedings of International Conference on Computer Vision Workshop (ICCVW)}, \bibinfo{year}{2011}.
\bibitem[{Scharstein and Pal(2007)}]{scharstein2007learning}
\bibinfo{author}{D.~Scharstein}, \bibinfo{author}{C.~Pal},
\newblock \bibinfo{title}{Learning conditional random fields for stereo},
\newblock in: \bibinfo{booktitle}{Proceedings of IEEE Conference on Computer Vision and Pattern Recognition (CVPR)}, \bibinfo{year}{2007}.
\bibitem[{Hirschmuller and Scharstein(2007)}]{hirschmuller2007evaluation}
\bibinfo{author}{H.~Hirschmuller}, \bibinfo{author}{D.~Scharstein},
\newblock \bibinfo{title}{Evaluation of cost functions for stereo matching},
\newblock in: \bibinfo{booktitle}{Proceedings of IEEE Conference on Computer Vision and Pattern Recognition (CVPR)}, \bibinfo{year}{2007}.
\bibitem[{Scharstein et~al.(2014)Scharstein, Hirschm{\"u}ller, Kitajima, Krathwohl, Ne{\v{s}}i{\'c}, Wang, and Westling}]{scharstein2014high}
\bibinfo{author}{D.~Scharstein}, \bibinfo{author}{H.~Hirschm{\"u}ller}, \bibinfo{author}{Y.~Kitajima}, \bibinfo{author}{G.~Krathwohl}, \bibinfo{author}{N.~Ne{\v{s}}i{\'c}}, \bibinfo{author}{X.~Wang}, \bibinfo{author}{P.~Westling},
\newblock \bibinfo{title}{High-resolution stereo datasets with subpixel-accurate ground truth},
\newblock in: \bibinfo{booktitle}{German conference on pattern recognition (GCPR)}, \bibinfo{organization}{Springer}, \bibinfo{year}{2014}, pp. \bibinfo{pages}{31--42}.
\bibitem[{Kim et~al.(2021)Kim, Ponce, and Ham}]{kim2021deformable}
\bibinfo{author}{B.~Kim}, \bibinfo{author}{J.~Ponce}, \bibinfo{author}{B.~Ham},
\newblock \bibinfo{title}{Deformable kernel networks for joint image filtering},
\newblock \bibinfo{journal}{International Journal on Computer Vision (IJCV)} \bibinfo{volume}{129} (\bibinfo{year}{2021}) \bibinfo{pages}{579--600}.
\bibitem[{He et~al.(2021)He, Zhu, Li, Bai, Cong, Zhang, Lin, Liu, and Zhao}]{he2021towards}
\bibinfo{author}{L.~He}, \bibinfo{author}{H.~Zhu}, \bibinfo{author}{F.~Li}, \bibinfo{author}{H.~Bai}, \bibinfo{author}{R.~Cong}, \bibinfo{author}{C.~Zhang}, \bibinfo{author}{C.~Lin}, \bibinfo{author}{M.~Liu}, \bibinfo{author}{Y.~Zhao},
\newblock \bibinfo{title}{Towards fast and accurate real-world depth super-resolution: Benchmark dataset and baseline},
\newblock in: \bibinfo{booktitle}{Proceedings of IEEE Conference on Computer Vision and Pattern Recognition (CVPR)}, \bibinfo{year}{2021}.
\bibitem[{Lutio et~al.(2019)Lutio, D'aronco, Wegner, and Schindler}]{lutio2019guided}
\bibinfo{author}{R.~d. Lutio}, \bibinfo{author}{S.~D'aronco}, \bibinfo{author}{J.~D. Wegner}, \bibinfo{author}{K.~Schindler},
\newblock \bibinfo{title}{Guided super-resolution as pixel-to-pixel transformation},
\newblock in: \bibinfo{booktitle}{Proceedings of International Conference on Computer Vision (ICCV)}, \bibinfo{year}{2019}.
\bibitem[{M{\"a}kinen et~al.(2019)M{\"a}kinen, Azzari, and Foi}]{makinen2019exact}
\bibinfo{author}{Y.~M{\"a}kinen}, \bibinfo{author}{L.~Azzari}, \bibinfo{author}{A.~Foi},
\newblock \bibinfo{title}{Exact transform-domain noise variance for collaborative filtering of stationary correlated noise},
\newblock in: \bibinfo{booktitle}{Proceedings of International Conference on Image Processing (ICIP)}, \bibinfo{year}{2019}.
\bibitem[{Krull et~al.(2019)Krull, Buchholz, and Jug}]{krull2019noise2void}
\bibinfo{author}{A.~Krull}, \bibinfo{author}{T.-O. Buchholz}, \bibinfo{author}{F.~Jug},
\newblock \bibinfo{title}{Noise2void-learning denoising from single noisy images},
\newblock in: \bibinfo{booktitle}{Proceedings of IEEE Conference on Computer Vision and Pattern Recognition (CVPR)}, \bibinfo{year}{2019}.
\bibitem[{Moran et~al.(2020)Moran, Schmidt, Zhong, and Coady}]{moran2020noisier2noise}
\bibinfo{author}{N.~Moran}, \bibinfo{author}{D.~Schmidt}, \bibinfo{author}{Y.~Zhong}, \bibinfo{author}{P.~Coady},
\newblock \bibinfo{title}{Noisier2noise: Learning to denoise from unpaired noisy data},
\newblock in: \bibinfo{booktitle}{Proceedings of IEEE Conference on Computer Vision and Pattern Recognition (CVPR)}, \bibinfo{year}{2020}.
\bibitem[{Wu et~al.(2020)Wu, Liu, Cao, Ren, and Zuo}]{wu2020unpaired}
\bibinfo{author}{X.~Wu}, \bibinfo{author}{M.~Liu}, \bibinfo{author}{Y.~Cao}, \bibinfo{author}{D.~Ren}, \bibinfo{author}{W.~Zuo},
\newblock \bibinfo{title}{Unpaired learning of deep image denoising},
\newblock in: \bibinfo{booktitle}{Proceedings of European Conference on Computer Vision (ECCV)}, \bibinfo{year}{2020}.
\bibitem[{Lehtinen et~al.(2018)Lehtinen, Munkberg, Hasselgren, Laine, Karras, Aittala, and Aila}]{lehtinen2018noise2noise}
\bibinfo{author}{J.~Lehtinen}, \bibinfo{author}{J.~Munkberg}, \bibinfo{author}{J.~Hasselgren}, \bibinfo{author}{S.~Laine}, \bibinfo{author}{T.~Karras}, \bibinfo{author}{M.~Aittala}, \bibinfo{author}{T.~Aila}, \bibinfo{title}{Noise2noise: Learning image restoration without clean data}, \bibinfo{year}{2018}.
\bibitem[{Kodak(1993)}]{kodak1993kodak}
\bibinfo{author}{E.~Kodak},
\newblock \bibinfo{title}{Kodak lossless true color image suite (photocd pcd0992)},
\newblock \bibinfo{journal}{Journal of Signal and Information Processing} \bibinfo{volume}{6} (\bibinfo{year}{1993}).
\bibitem[{Movahedi and Elder(2010)}]{movahedi2010design}
\bibinfo{author}{V.~Movahedi}, \bibinfo{author}{J.~H. Elder},
\newblock \bibinfo{title}{Design and perceptual validation of performance measures for salient object segmentation},
\newblock in: \bibinfo{booktitle}{Proceedings of IEEE Conference on Computer Vision and Pattern Recognition Workshop (CVPRW)}, \bibinfo{year}{2010}.
\bibitem[{Martin et~al.(2001)Martin, Fowlkes, Tal, and Malik}]{martin2001database}
\bibinfo{author}{D.~Martin}, \bibinfo{author}{C.~Fowlkes}, \bibinfo{author}{D.~Tal}, \bibinfo{author}{J.~Malik},
\newblock \bibinfo{title}{A database of human segmented natural images and its application to evaluating segmentation algorithms and measuring ecological statistics},
\newblock in: \bibinfo{booktitle}{Proceedings of International Conference on Computer Vision (ICCV)}, \bibinfo{year}{2001}.
\bibitem[{Gao et~al.(2022)Gao, Li, Li, Wu, Lu, and Yu}]{gao2022feature}
\bibinfo{author}{G.~Gao}, \bibinfo{author}{W.~Li}, \bibinfo{author}{J.~Li}, \bibinfo{author}{F.~Wu}, \bibinfo{author}{H.~Lu}, \bibinfo{author}{Y.~Yu},
\newblock \bibinfo{title}{Feature distillation interaction weighting network for lightweight image super-resolution},
\newblock in: \bibinfo{booktitle}{Proceedings of the AAAI conference on artificial intelligence}, volume~\bibinfo{volume}{36}, \bibinfo{year}{2022}, pp. \bibinfo{pages}{661--669}.
\bibitem[{Wang et~al.(2022)Wang, Wan, Yang, Li, Chau, and Kot}]{wang2022low}
\bibinfo{author}{Y.~Wang}, \bibinfo{author}{R.~Wan}, \bibinfo{author}{W.~Yang}, \bibinfo{author}{H.~Li}, \bibinfo{author}{L.-P. Chau}, \bibinfo{author}{A.~Kot},
\newblock \bibinfo{title}{Low-light image enhancement with normalizing flow},
\newblock in: \bibinfo{booktitle}{Proceedings of the AAAI Conference on Artificial Intelligence (AAAI)}, \bibinfo{year}{2022}.
\bibitem[{Wei et~al.(2018)Wei, Wang, Yang, and Liu}]{wei2018deep}
\bibinfo{author}{C.~Wei}, \bibinfo{author}{W.~Wang}, \bibinfo{author}{W.~Yang}, \bibinfo{author}{J.~Liu}, \bibinfo{title}{Deep retinex decomposition for low-light enhancement}, \bibinfo{year}{2018}.
\bibitem[{Liu et~al.(2021)Liu, Xu, Yang, Fan, and Huang}]{liu2021benchmarking}
\bibinfo{author}{J.~Liu}, \bibinfo{author}{D.~Xu}, \bibinfo{author}{W.~Yang}, \bibinfo{author}{M.~Fan}, \bibinfo{author}{H.~Huang},
\newblock \bibinfo{title}{Benchmarking low-light image enhancement and beyond},
\newblock \bibinfo{journal}{International Journal of Computer Vision} \bibinfo{volume}{129} (\bibinfo{year}{2021}) \bibinfo{pages}{1153--1184}.
\bibitem[{Bevilacqua et~al.(2012)Bevilacqua, Roumy, Guillemot, and Alberi-Morel}]{bevilacqua2012low}
\bibinfo{author}{M.~Bevilacqua}, \bibinfo{author}{A.~Roumy}, \bibinfo{author}{C.~Guillemot}, \bibinfo{author}{M.~L. Alberi-Morel},
\newblock \bibinfo{title}{Low-complexity single-image super-resolution based on nonnegative neighbor embedding},
\newblock in: \bibinfo{booktitle}{Proceedings of British Machine Vision Conference (BMVC)}, \bibinfo{year}{2012}.
\bibitem[{Zeyde et~al.(2012)Zeyde, Elad, and Protter}]{zeyde2012single}
\bibinfo{author}{R.~Zeyde}, \bibinfo{author}{M.~Elad}, \bibinfo{author}{M.~Protter},
\newblock \bibinfo{title}{On single image scale-up using sparse-representations},
\newblock in: \bibinfo{booktitle}{Curves and Surfaces: 7th International Conference, Avignon, France, June 24-30, 2010, Revised Selected Papers 7}, \bibinfo{year}{2012}.
\bibitem[{Matsui et~al.(2017)Matsui, Ito, Aramaki, Fujimoto, Ogawa, Yamasaki, and Aizawa}]{matsui2017sketch}
\bibinfo{author}{Y.~Matsui}, \bibinfo{author}{K.~Ito}, \bibinfo{author}{Y.~Aramaki}, \bibinfo{author}{A.~Fujimoto}, \bibinfo{author}{T.~Ogawa}, \bibinfo{author}{T.~Yamasaki}, \bibinfo{author}{K.~Aizawa},
\newblock \bibinfo{title}{Sketch-based manga retrieval using manga109 dataset},
\newblock \bibinfo{journal}{Multimedia Tools and Applications} \bibinfo{volume}{76} (\bibinfo{year}{2017}) \bibinfo{pages}{21811--21838}.
\bibitem[{Dong et~al.(2022)Dong, Cao, and Fu}]{dong2022incremental}
\bibinfo{author}{Q.~Dong}, \bibinfo{author}{C.~Cao}, \bibinfo{author}{Y.~Fu},
\newblock \bibinfo{title}{Incremental transformer structure enhanced image inpainting with masking positional encoding},
\newblock in: \bibinfo{booktitle}{Proceedings of IEEE Conference on Computer Vision and Pattern Recognition (CVPR)}, \bibinfo{year}{2022}.
\bibitem[{Nazeri et~al.(2019)Nazeri, Thasarathan, and Ebrahimi}]{nazeri2019edge}
\bibinfo{author}{K.~Nazeri}, \bibinfo{author}{H.~Thasarathan}, \bibinfo{author}{M.~Ebrahimi},
\newblock \bibinfo{title}{Edge-informed single image super-resolution},
\newblock in: \bibinfo{booktitle}{Proceedings of International Conference on Computer Vision Workshop (ICCVW)}, \bibinfo{year}{2019}.
\bibitem[{Chen et~al.(2022)Chen, Chu, Zhang, and Sun}]{chen2022simple}
\bibinfo{author}{L.~Chen}, \bibinfo{author}{X.~Chu}, \bibinfo{author}{X.~Zhang}, \bibinfo{author}{J.~Sun},
\newblock \bibinfo{title}{Simple baselines for image restoration},
\newblock in: \bibinfo{booktitle}{Computer Vision--ECCV 2022: 17th European Conference, Tel Aviv, Israel, October 23--27, 2022, Proceedings, Part VII}, \bibinfo{organization}{Springer}, \bibinfo{year}{2022}, pp. \bibinfo{pages}{17--33}.
\bibitem[{Dauphin et~al.(2017)Dauphin, Fan, Auli, and Grangier}]{dauphin2017language}
\bibinfo{author}{Y.~N. Dauphin}, \bibinfo{author}{A.~Fan}, \bibinfo{author}{M.~Auli}, \bibinfo{author}{D.~Grangier},
\newblock \bibinfo{title}{Language modeling with gated convolutional networks},
\newblock in: \bibinfo{booktitle}{International conference on machine learning}, \bibinfo{organization}{PMLR}, \bibinfo{year}{2017}, pp. \bibinfo{pages}{933--941}.
\bibitem[{Alhashim and Wonka(2018)}]{alhashim2018high}
\bibinfo{author}{I.~Alhashim}, \bibinfo{author}{P.~Wonka},
\newblock \bibinfo{title}{High quality monocular depth estimation via transfer learning},
\newblock \bibinfo{journal}{arXiv preprint arXiv:1812.11941}  (\bibinfo{year}{2018}).
\bibitem[{Sandler et~al.(2018)Sandler, Howard, Zhu, Zhmoginov, and Chen}]{sandler2018mobilenetv2}
\bibinfo{author}{M.~Sandler}, \bibinfo{author}{A.~Howard}, \bibinfo{author}{M.~Zhu}, \bibinfo{author}{A.~Zhmoginov}, \bibinfo{author}{L.-C. Chen},
\newblock \bibinfo{title}{Mobilenetv2: Inverted residuals and linear bottlenecks},
\newblock in: \bibinfo{booktitle}{Proceedings of the IEEE conference on computer vision and pattern recognition}, \bibinfo{year}{2018}, pp. \bibinfo{pages}{4510--4520}.
\bibitem[{Nvidia(2023{\natexlab{a}})}]{nvidiaorin}
\bibinfo{author}{Nvidia}, \bibinfo{title}{{Jetson Orin Developer Kit}}, \bibinfo{howpublished}{\url{https://www.nvidia.com/en-us/autonomous-machines/embedded-systems/jetson-orin/}}, \bibinfo{year}{2023}{\natexlab{a}}.
\bibitem[{Nvidia(2023{\natexlab{b}})}]{nvidiatensorrt}
\bibinfo{author}{Nvidia}, \bibinfo{title}{{NVIDIA TensorRT}}, \bibinfo{howpublished}{\url{https://developer.nvidia.com/tensorrt}}, \bibinfo{year}{2023}{\natexlab{b}}.
\bibitem[{Logitech(2023)}]{webcam}
\bibinfo{author}{Logitech}, \bibinfo{title}{{Logitech C920 PRO HD Webcam}}, \bibinfo{howpublished}{\url{https://www.logitech.com/en-eu/products/webcams/c920-pro-hd-webcam.960-001055.html}}, \bibinfo{year}{2023}.
\bibitem[{Silberman et~al.(2012)Silberman, Hoiem, Kohli, and Fergus}]{silberman2012indoor}
\bibinfo{author}{N.~Silberman}, \bibinfo{author}{D.~Hoiem}, \bibinfo{author}{P.~Kohli}, \bibinfo{author}{R.~Fergus},
\newblock \bibinfo{title}{Indoor segmentation and support inference from rgbd images},
\newblock in: \bibinfo{booktitle}{Proceedings of European Conference on Computer Vision (ECCV)}, \bibinfo{year}{2012}.

\end{thebibliography}
\small

\newpage

\section*{Author Biography}
\noindent \textbf{Jisu Shin} received the B.S. degree from the Department of Electronic Engineering and Computer Science, Gwangju Institute of Science and Technology, Gwangju, South Korea, in 2021. She is currently working toward the PhD degree in the Artificial Intelligence Graduate School, Gwangju Institute of Science and Technology, Gwangju, South Korea. Her current research includes scene reconstruction on both 2D and 3D and human reconstruction.
\subsection*{  } 
\noindent \textbf{Seunghyun Shin}
received the B.S. degree from the Department of Electronic Engineering and Computer Science, Gwangju Institute of Science and Technology, Gwangju, South Korea, in 2021. She is currently working toward the PhD degree in the Artificial Intelligence Graduate School, Gwangju Institute of Science and Technology, Gwangju, South Korea. His current research includes computer vision, especially low-level vision tasks and aesthetic image enhancement.
\subsection*{  } 
\noindent \textbf{Hae-Gon Jeon}
received the B.S. degree from the School of Electrical and Electronic Engineering, Yonsei University, in 2011, and the MS and PhD degrees from the School of Electrical Engineering, KAIST, in 2013 and 2018, respectively. He is currently working at the Artificial Intelligence Graduate School of GIST as an associate professor. He is a winner of the Best PhD Thesis Award in 2018 at KAIST. His research interests include computational imaging, 3D reconstruction and machine learning. He is a member of the IEEE.




\end{document}